\documentclass[manuscript,screen]{acmart}
\AtBeginDocument{%
  }

\setcopyright{acmlicensed}
\copyrightyear{2024}
\acmYear{2024}
\acmDOI{XXXXXXX.XXXXXXX}

\acmISBN{978-1-4503-XXXX-X/18/06}

\usepackage{longtable}

\usepackage{makecell}
\usepackage{soul}
\usepackage{multirow}
\usepackage{rotating}
\usepackage{lscape}

\newcommand{\fakepar}[1]{\noindent\textbf{#1:}}

\newcommand{\fakesubpar}[1]{\noindent\textit{#1:}}

\begin{document}

\title{On-device Training: A First Overview on Existing Systems}

\author{Shuai Zhu}
\orcid{0000-0002-9839-3820}
\affiliation{%
  \institution{RISE Research Institutes of Sweden \& Uppsala University}
  \city{Stockholm}
  \country{Sweden}}
\email{shuai.zhu@ri.se}

\author{Thiemo Voigt}
\affiliation{%
  \institution{Uppsala University \& RISE Research Institutes of Sweden}
  \city{Uppsala \& Stockholm}
  \country{Sweden}
}
\email{thiemo.voigt@angstrom.uu.se}

\author{Fatemeh Rahimian}
\affiliation{%
  \institution{RISE Research Institutes of Sweden}
  \city{Stockholm}
  \country{Sweden}
}
\email{fatemeh.rahimian@ri.se}

\author{JeongGil Ko}
\orcid{0000-0003-0799-4039}
\affiliation{%
  \institution{Yonsei University \& POSTECH}
  \city{Seoul \& Pohang}
  \country{South Korea}
}
\email{jeonggil.ko@yonsei.ac.kr (Corresponding Author)}
\renewcommand{\shortauthors}{Zhu et al.}

\begin{abstract}
   The recent breakthroughs in machine learning (ML) and deep learning (DL) have catalyzed the design and development of various intelligent systems over wide application domains. While most existing machine learning models require large memory and computing power, efforts have been made to deploy some models on resource-constrained devices as well. A majority of the early application systems focused on exploiting the inference capabilities of ML and DL models, where data captured from different mobile and embedded sensing components are processed through these models for application goals such as classification and segmentation. More recently, the concept of exploiting the mobile and embedded computing resources for ML/DL model training has gained attention, as such capabilities allow (i) the training of models via local data without the need to share data over wireless links, thus enabling privacy-preserving computation by design, (ii) model personalization and environment adaptation, and (iii) deployment of accurate models in remote and hardly accessible locations without stable internet connectivity. This work summarizes and analyzes state-of-the-art systems research that allows such on-device model training capabilities and provides a survey of on-device training from a systems perspective.
\end{abstract}

\ccsdesc[500]{General and reference~Surveys and overviews}
\ccsdesc[500]{Computing methodologies~Machine learning}
\ccsdesc[500]{Computing methodologies~Learning paradigms}
\ccsdesc[500]{Hardware~Sensor devices and platforms}

\keywords{Machine Learning, IoT devices, On-device Training}


\maketitle

\section{Introduction}
\label{Sec:Introduction}
Machine learning (ML) has been extensively studied in recent years and has matured enough to be practically integrated into real-world computing systems. Data-driven ML-based system design has catalyzed the development of countless applications in many application domains, including those targeted by Internet of Things (IoT) systems. Concurrently, with the fast-growing interest and advancements in technology that support IoT systems, the number of IoT devices deployed in our everyday environments is significantly increasing. GSMA Intelligence~\footnote{\url{https://www.gsmaintelligence.com/}} forecasts that the number of IoT devices will increase to 25 billion by 2025. IoT devices with various sensing components are being ubiquitously adopted, taking the role of capturing data for various application scenarios, including healthcare, surveillance, and environmental monitoring. Data generated from these devices can provide opportunities for applying data-driven ML methods to accomplish their application goals. Advances in ML, including Deep Learning (DL), have resulted in tremendous improvements in solving various types of problems in computer vision, natural language processing, machine translation, etc., and across different application domains such as biology and healthcare, automotive industries, smart cities, and many more. To fully leverage the advances in both IoT and ML, on-device ML integrates ML algorithms on IoT devices and is becoming an active area of research.

Largely, on-device ML consists of on-device inference and training operations. On-device inference refers to deploying pre-trained ML models on devices and locally performing inference operations such as classification or regression. Researchers have presented extensive works using such an `inference-embedded' system architecture~\cite{Comprehensive_Benchmark,nemo,DeepCache,appqueryseroniotcamera,DeepWear,alookatdnnonsamrtphone,asymo,spinn,ulayer,nestDNN}. On the other hand, on-device training takes a step further and targets to \textit{train} the models locally. Note that, typically, due to computational power limitations, cloud-based (or server-based training) is the mainstream approach, where local data is shared with the server, and the training operations execute at remote platforms. Later, updated models can be re-distributed to local IoT devices to perform local inference operations. However, since holding the capability to locally train an ML model can preserve the precious network bandwidth and limited battery budgets and at the same time contain the raw data locally to preserve privacy, researchers have recently started to propose schemes to train ML models on-device, despite their computational resource limitations~\cite{patil2022poet, Mandheling, minilearn, Melon, Sage, ren2021tinyol, cai2020tinytl, lin2022device}. Still, these works are in their early stages, and we believe that a comprehensive view of what is done and what is left, can catalyze further research ahead. Unlike previous work that analyzes the research space from an algorithmic perspective~\cite{dhar2021survey}, we take a systematic view and present a survey on systems and frameworks for supporting on-device training. On-device training comes with the following benefits: 
\begin{itemize}
\item 
From the device perspective, the main advantages are that on-device training can take place even in the absence of an Internet connection. Secondly,  there is no need to upload data to the cloud and/or download an updated model back, and that in itself saves bandwidth and reduces latency and energy. The latter is important for low-power IoT devices, where communication is typically far more expensive than computing. It is also important when dealing with time-critical applications that should react quickly without communicating with a server or over a network.
\item 
From a data perspective, training on local devices is privacy-preserving by design. Privacy is a critical issue in many real-world use cases and has slowed down the deployment of AI models in such cases~\cite{dwork2014algorithmic}. Enabling on-device training removes this obstacle and enables the utilization of ML models in reality. 
\item 
From a modeling perspective,  devices can become smarter as they can cope with model drift problems~\cite{modeldrift} and retrain the deployed pre-trained models in order to adapt to the environment and even the end-user. For example, a medical device can over time learn to provide personalized predictions or services that fit the specific conditions of a certain patient.
\end{itemize}

\subsection{Challenges of On-device Training}
On-device training entails the following three main technical challenges.
\begin{itemize}
\item
\fakepar{Mismatch between hardware resources availability and demand}  IoT devices typically own limited hardware resources, for example, the memory capacity is in the order of kilobytes or megabytes. However, ML model training requires considerable hardware, including computing and memory resources. Concretely, model training could be described as an optimization problem. The optimization goal is to minimize the loss, which measures the difference between the output of the model and the ground truth. During the training process, ML methods keep tuning model parameters by following the gradient direction. Gradient-based training normally applies backpropagation algorithms to compute the gradient. The backpropagation algorithm stores a large amount of intermediate loss and output of hidden layers, which could consume memory on the order of tens of gigabytes or more. This contradiction makes on-device training challenging. 
\item
\fakepar{High heterogeneity of IoT devices} IoT devices are diverse and heterogeneous. Specifically, apart from the heterogeneity in vendors, IoT devices cover a wide spectrum in terms of their capability: from severely constrained microcontrollers, such as Nordic Semiconductor nRF9160~\footnote{https://www.nordicsemi.com/products/nrf9160}, to single-board computers with more capable hardware, such as Raspberry Pi 4~\footnote{https://www.raspberrypi.com/products/raspberry-pi-4-model-b/}. Therefore, it is challenging to propose a single and generally applicable solution for all devices.
\item
\fakepar{Limited existing work} Most existing model training optimization algorithms, such as batch normalization (BN)~\cite{IoffeS15BN} and Adam~\cite{kingma2014adam},  either target to achieve higher accuracy or converge faster. These works often overlook the resource constraints of edge devices, such as limited memory and battery power.
\end{itemize}

\subsection{Scope of this Survey}
\label{subsec:scope}
This survey summarizes current state-of-the-art (SOTA)  systems research on on-device training. Such systems aim to enable neural network training on resource-constraint devices, ranging from modern mobile platforms to microcontrollers. 
Our survey also includes modern smartphones but does not include the NVIDIA Jetson platform~\cite{nvidia2022jetson}. Some modern smartphones are equipped with dedicated accelerators, such as GPUs and NPUs. However, smartphones are user-centric and integrate multiple functions, such as communication, entertainment, and photography, which share hardware resources.  Excessive resource consumption by one application can lead to degradation of the user experience. In addition,  smartphones are powered by built-in batteries, so applications also need to consider power consumption. Systems designed for smartphones must contend with various resource constraints, although smartphones appear to involve abundant hardware resources. Therefore,  this survey includes systems designed for smartphones. On the other hand, the NVIDIA Jetson platform is a powerful tool for specialized artificial intelligence and robotics applications. Hence, this survey will not provide a detailed analysis of systems that are only evaluated on NVIDIA Jetson platforms. In addition, this work focuses on single-device training and does not include the body of work in collaborative model training (e.g., federated learning). 

The articles we survey are published in renowned conferences on relevant topics, including mobile computing, wireless and mobile networking, and ML. Specifically, this survey focuses on recently published papers (2019-2023) from MobiCom, SenSys, IPSN, MobiSys, and EWSN, as well as AAAI, ICLR, ICML, NeurIPS, and IJCAI. Apart from these conferences, we also use Google Scholar and Microsoft Research to identify other relevant work. Keywords used to obtain these papers are {\tt on-device/edge/device + learning/training/adaptation/update}. Overall, we choose and direct our focus to analyzing eleven representative systems.

\subsection{Outline}
The rest of the paper is organized as follows. Section~\ref{Sec:systems} provides an overview of current on-device training systems and the target devices of these systems. We present an analysis from the Machine Learning perspective in Section~\ref{Sec:ml} and Section~\ref{Sec:optimaztiontechs} presents optimization techniques to enable on-device model training. Section~\ref{Sec:result} discusses the results reported from current SOTA systems. Before presenting the conclusions, Section~\ref{sec:discussion} elaborates on the methods and ideas presented in the current SOTA system works for on-device training. Finally, Section~\ref{sec:conclusion} discusses potential future directions and concludes this article.

\section{Overview of Existing Systems}
\label{Sec:systems}
We survey the following systems for on-device training: POET~\cite{patil2022poet}, Mandheling~\cite{Mandheling}, MiniLearn~\cite{minilearn}, Melon~\cite{Melon}, Sage~\cite{Sage}, TinyTL~\cite{cai2020tinytl}, TTE~\cite{lin2022device}, ElasticTrainer~\cite{huang2023elastictrainer}, MDLdroidLite~\cite{zhang2020mdldroidlite}, LifeLearner~\cite{kwon2023lifelearner}, and zTT~\cite{kim2021ztt}. In summary, existing systems operate within four paradigms: one-stage, two-stage, continual learning, 
and reinforcement learning paradigms. Firstly, the one-stage paradigm systems apply various optimization techniques, such as pruning~\cite{molchanov2016pruning} and quantization~\cite{jacob2018quantization}, to perform model training. Secondly, unlike the one-stage paradigm, the two-stage paradigm has an additional preparation stage, where systems prepare the computing graph and generate the training plan before performing model training. The plan schedules when and how to use specific techniques during training. Thirdly, continual learning-based systems enable Deep Neural Network (DNN) models to learn new tasks throughout their lifetime. 
MDLdroidLite and LifeLearner are two continual learning-based systems that apply continuous growth and memory replay techniques to facilitate on-device training, respectively. Details are introduced in Section~\ref{subsec:MDLdroidLite}. Another system, Miro~\cite{ma2023cost}, also applies continual learning to enable on-device training. However, the authors only evaluate the feasibility of the system on NVIDIA Jetson platforms (NVIDIA Jetson TX2 and NVIDIA Jetson Xavier NX). Therefore, this survey only provides a brief summary of Miro.
Fourthly,  zTT employs reinforcement learning to enable on-device training. The system focuses on learning-based Dynamic Voltage and Frequency Scaling (DVFS) without thermal throttling for mobile devices to dynamically adjust CPU and GPU voltage-frequency levels during runtime, which effectively manages heat generation and power consumption. This approach aims to strike a balance between energy efficiency and performance. 

This section begins by providing an overview of relevant review papers on on-device training. It then delves into the distinction between on-device training and general acceleration techniques for DNN training. Furthermore, it provides an overview of the devices targeted by existing systems. Finally, the section outlines the key design choices adopted by these systems.

\subsection{Relevant Reviews on On-Device Training}
This paper provides a comprehensive overview of on-device training from a \textit{system perspective}, focusing on enabling model training on single resource-constrained devices. While Zhao et al.\cite{zhao2022survey} offer a summary and comparison of various DL applications on mobile devices, emphasizing on-device inference for tasks like computer vision, speech/speaker recognition, and transportation mode detection, they only touch briefly on training methodologies, particularly federated learning. Similarly, Saha et al.\cite{saha2022machine} concentrate on techniques for ML inference on microcontroller-class devices, discussing features like model pruning, quantization, and lightweight neural network architecture design. Their discussions provide a high-level overview of training methods, including last-layer transfer learning, quantized continual learning, and training specialized operators (e.g., TinyTL~\cite{cai2020tinytl}). In contrast, our work provides a more detailed summary and analysis of on-device training techniques, offering insights into how these methods enable efficient model training directly on resource-constrained devices.

In addition, Dhar et al.~\cite{dhar2021survey} provides a comprehensive review of on-device training from an algorithms and theory perspective. Unlike the focus on system-level considerations in our paper, Dhar et al. delve into the algorithmic and theoretical aspects of on-device training. They categorize SOTA designs into five approaches, including lightweight ML algorithms, model complexity reduction, optimization routine modifications (e.g., quantization), data compression, and new protocols for data observation. Furthermore, the authors offer insights into theoretical considerations, urging researchers to analyze resource constraints, assess applicability, focus on algorithm development, and extend existing theories. To the best of our knowledge, this represents the first review of on-device training with a focus on algorithms and theory.

\subsection{Comparison with General Speed-up Techniques for DNN Training}
\label{subsec:generaltech}

The efficiency of DNN training is a critical area of research, given the extensive hardware resources, both memory and computation resources, required for such tasks. This section discusses the prevalent techniques for high-efficiency DNN training and the distinctions between them and on-device training. Specifically, the efficiency metrics for DL consist of two representative types:  computation-related and memory-related metrics. Hence, general speeding-up techniques for training aim to optimize either one or both of these. Concretely, these approaches aim to reduce training time and resource usage while maintaining (or even improving) model performance. In contrast, while on-device training systems generally share similar rationale and goals, they face an additional layer of constraints, including energy resources, memory, and computation resources constraints, which pose additional challenges for on-device training.

\subsubsection{Computational Efficiency Optimization Approaches} Optimizing computational efficiency for DL model training is crucial for speeding up experiments, reducing costs, and enabling the training of larger models. Those approaches generally optimize computation in three aspects: hardware, data, and algorithms, which are the three pillars of DL~\cite{goodfellow2016deep}.

\fakepar{Hardware acceleration} Hardware acceleration techniques aim to design specialized hardware units to speed-up the computation time for both training and inference operations. The most common hardware accelerators are graphics processing units (GPUs)~\cite{mishra2021accelerating} and tensor processing units (TPUs)~\cite{jouppi2017datacenter}. For example, the NVIDIA H100 Tensor Core GPU~\footnote{https://www.nvidia.com/en-us/data-center/h100/} can accelerate exascale workloads. However, hardware accelerators for edge devices are significantly less powerful than accelerators in the server/cloud, and many processing components, such as microcontrollers, are not equipped with and do not support dedicated hardware accelerators. In addition, platforms with limited energy resources, such as those that are battery-powered, cannot afford to perform highly parallel computations with accelerators. In contrast, on-device training systems enable highly resource-efficient on-device training by leveraging other types of assisting hardware, such as Digital Signal Processors (DSPs)~\cite{Mandheling}. Overall, while both general speed-up techniques for DNN training and on-device training leverage assisting hardware, they focus on different performance aspects. General techniques mainly concern the training speed, while on-device training systems aim to \textit{complete} the training with constrained computing resources.

\fakepar{Data optimization and management} Data optimization and management for DNN training aims to optimize the data pipeline and task processing between memory units and computing units to enhance computing performance and efficiency by exploiting techniques such as data preprocessing, caching strategies, data pipeline optimization for I/O bottleneck minimization, and datasets organization to facilitate easy access and manipulation. In this regard, on-device training and general speed-up techniques share similar goals. 

\fakepar{Algorithm efficiency} Algorithm efficiency in DNN model training refers to how effectively a given algorithm uses computational resources, such as GPUs, to achieve training objectives within a reasonable amount of time. One typical general speed-up technique is to leverage parallelism and distributed strategies for training. For example, SALIENT~\cite{kaler2022accelerating} achieves a speed-up of $3\times$ over a standard PyTorch-Geometric~\cite{fey2019fast} implementation with a single GPU and a further $8\times$ parallel speed-up with 16 GPUs. 
The goal is to achieve high accuracy and consume less computational resources. For instance,  Ekya~\cite{bhardwaj2022ekya} is a continuous learning system for video analytics models on edge servers. The system solves the challenge of jointly supporting inference and retraining tasks, which trades-off the accuracy of the retrained model with the inference accuracy. Compared to the baseline~\cite{zhang2017live}, Ekya achieves 29\% higher accuracy, and to reach the same accuracy as Ekya, the baseline requires $4\times$ more GPU resources. Unfortunately, on-device training systems do not have the luxury of utilizing such powerful hardware resources. Overall, general speed-up techniques aim to accelerate the training process and achieve start-of-the-art performance, while the goal of many on-device training schemes is to enable training and achieve reasonably good performance.

\subsubsection{Memory Reduction Approaches} As DL research has evolved towards employing deeper models and larger mini-batches to pursue higher accuracy, the memory blowup issue in DNN training has become a significant research problem. Specifically, the memory usage of training operations can be $5\times$ to $100\times$ higher compared to inference depending on the input and batch sizes~\cite{Sage}. 
Even for server-scale GPUs, the increasing memory demand in model training becomes challenging to accommodate. To bridge the gap between memory demand and supply, researchers have made various efforts to reduce model training memory overhead. Specifically, memory reduction approaches can be broadly classified in five categories~\cite{Sage}: (1) gradient accumulation, (2) gradient checkpointing, (3) reducing activation size, (4) reducing activation count, and (5) memory virtualization. Concretely, the first four categories pertain to software- and algorithm-level memory usage optimization for DNN training speed-up. These approaches are applicable to both general training speed-up and on-device training. On the other hand, the fifth class leverages memory virtualization to speed up training. Specifically, this approach simultaneously utilizes both GPU and CPU memory for training speed-up. However, differing from servers where the CPU and GPU are paired with separate hardware memory, most mobile System-on-Chips (SoCs) adopt a unified memory architecture, which means different processors share a common physical memory. Furthermore, the mobile operating system, background workloads, and applications can cause unpredictable memory churn that leads to scarce and dynamic memory availability for DNN training. Therefore, memory virtualization approaches do not apply to on-device training. 

\fakepar{Gradient accumulation} This approach aims to optimize memory usage by adopting incremental computation techniques for mini-batch operations~\cite{stein2021latency}. The computation of gradients for an entire mini-batch is advantageous for maximizing hardware utilization through enhanced parallelism. However, platforms with limited resources may face challenges meeting the memory requirements for processing mini-batches in their entirety. In theory, mini-batches can be partitioned into smaller sub-mini-batches, referred to as micro-batches. In this approach, gradients for individual micro-batches are computed separately, and their results are aggregated to derive the gradient for the complete mini-batch. A critical aspect lies in determining the appropriate size for splitting the mini-batch, as excessively small batch sizes can result in under-utilization of hardware resources.

\fakepar{Gradient checkpointing}  This method selectively retains a subset of activations in memory while recomputing demanded activations that are not stored~\cite{Checkpointing}. This strategy involves a trade-off between memory utilization and additional computation. The key challenge lies in effectively determining which activations to store and which to discard. Previous research efforts have explored computing optimal schedules to minimize computation overhead within a given hardware budget~\cite{kusumoto2019graph, jain2020checkmate}. More recently, dynamic gradient checkpointing has emerged as a heuristic approach to identifying on-memory activations with minimal computing overhead~\cite{kirisame2020dynamic}. Gradient checkpointing effectively reduces memory footprints across activation layers, while accumulation operates over mini-batches. Importantly, these approaches are orthogonal and can seamlessly coexist~\cite{Sage}.

\fakepar{Activation Size Reduction} This strategy is widely employed to mitigate the memory consumption of models and primarily involves quantization and sparsification techniques. Quantization methods utilize lower-precision floating-point or integer units for calculating and storing intermediate states, thereby reducing model operation costs~\cite{micikevicius2017mixed}. Additionally, these methods are hardware-friendly, as most GPUs, including mobile GPUs, support 16-bit floating-point or 8-bit integer arithmetic, which offers efficient computation compared to the commonly used 32-bit floating-point arithmetic. However, while sparsification approaches also reduce gradient size, current hardware has exhibited inefficiencies in storing and computing sparse matrices~\cite{gale2020sparse}.

\fakepar{Activation Number Reduction} Another viable approach for alleviating model computation memory is by reducing the number of intermediate states in a model. For example, researchers merge a series of arithmetic operations in a computational graph into a single composite operation to consolidate multiple gradients into a unified state. Such graph-level optimizations are employed for faster and memory-efficient inference~\cite{chen2018tvm, rotem2018glow}, although their application in training remains largely unexplored.

\fakepar{Memory Virtualization} This method leverages host-side memory in GPUs to logically extend usable memory. Several studies have employed such schemes for server-scale DNN training~\cite{huang2020swapadvisor, peng2020capuchin, rhu2016vdnn}. However, this approach is not suitable for mobile environments, as mobile SoCs typically employ a unified memory architecture for both the GPU and CPU.

\subsection{Selected Systems and their device categories}

The scope of this survey covers on-device training systems tailored for resource-constrained devices, with a focus on memory limitations. Specifically, we classify devices into three categories based on their hardware resources: high-constraint devices, middle-constraint devices, and low-constraint devices. High-constraint devices typically have hundreds of kilobytes of memory, while middle-constraint devices possess hundreds of megabytes. Low-constraint devices, on the other hand, have less than ten gigabytes of memory. Table~\ref{tab:device} outlines the hardware specifications of the devices utilized in previous studies, while Table~\ref{tab:targetdevice} provides an overview of the target devices for the systems research works we review. 

\begin{longtable}[h!]{|c|c|c|}
   \caption{The hardware feature of targeted devices. All information from corresponding official websites.}\label{tab:device}\\ \hline
   
 \textbf{Device} & \textbf{Memory} & \textbf{Computing Resource}\\ \hline
 \endfirsthead

 \hline
 \multicolumn{3}{|c|}{Continuation of Table~\ref{tab:device}}\\
 \hline
 \textbf{Device} & \textbf{Memory} & \textbf{Computing Resource}\\ 
 \endhead
 \endfoot

 \multicolumn{3}{| c |}{End of Table~\ref{tab:device}}\\
 \hline
 \endlastfoot
   
    NVIDIA Jetson TX2  & 8 GB RAM &  \makecell{Quad-Core ARM Cortex-A57 MPCore CPU \\ 256-core NVIDIA Pascal GPU} \\ \hline

    NVIDIA Jetson Nano  &  4 GB RAM &   \makecell{Quad-core ARM Cortex-A57 MPCore CPU \\ 128-core NVIDIA Pascal GPU} \\ \hline
    
    Samsung Galaxy S10 & 8 GB RAM & \makecell{Exynos 9820 SoC:\textit{ Octa-Core CPU with} \\ \textit{two Exynos M4, two Cortex-A75 and }\\ \textit{four Cortex-A55 cores, and Mali-G76 GPU}} \\ \hline 
    
    Samsung Note 20    & 8 GB RAM & \makecell{Snapdragon 865 SoC:\textit{ Octa-core Kryo 585 CPU, } \\ \textit{Adreno 650 GPU and Hexagon 698 DSP}}\\ \hline
    
    Samsung Note 10    & 8 GB RAM & \makecell{Snapdragon 855 SoC:\textit{ Octa-core Kryo 485 CPU, }\\ \textit{ Adreno 640 GPU and Hexagon 690 DSP}} \\ \hline
   
    Xiaomi 11 Pro      &  8 GB RAM & \makecell{Snapdragon 888 SoC:\textit{ Octa-core Kryo 680 CPU, } \\ \textit{Adreno 660 GPU and Hexagon 780 DSP}} \\ \hline
    
    Xiaomi 10          &  8 GB  RAM & \makecell{Snapdragon 865 SoC:\textit{ Octa-core Kryo 585 CPU, } \\ \textit{Adreno 650 GPU and Hexagon 698 DSP}} \\ \hline
    
    \makecell{Redmi Note 9 Pro \\(Mandheling~\cite{Mandheling})}  &  8GB RAM & \makecell{Snapdragon 750G SoC:\textit{ Octa-core CPU with}\\ \textit{two Kryo 570 cores and six Kryo 570} \\ \textit{cores, Adreno 619 GPU and Hexagon 694 DSP}}\\ \hline
    
    \makecell{Redmi Note 9 Pro \\ (Melon~\cite{Melon})}  & 6 GB RAM & \makecell{Snapdragon 720G SoC:\textit{ Octa-core CPU with} \\ \textit{two Kryo 260 Gold cores and six Kryo 260 Silver } \\ \textit{cores, Adreno 618 GPU and Hexagon 692 DSP}} \\ \hline

    Redmi Note 8       & 4 GB RAM & \makecell{Snapdragon 665 SoC:\textit{ Octa-core CPU with} \\ \textit{four Kryo 260 Gold cores and four Kryo 260 Silver } \\ \textit{cores, Adreno 610 GPU and Hexagon 686 DSP}} \\ \hline

    Meizu 16T          &  6GB RAM  &  \makecell{Snapdragon 855 SoC:\textit{ Octa-core Kryo 485 CPU, }\\ \textit{ Adreno 640 GPU and Hexagon 690 DSP}}  \\ \hline
    
    vivo iQOO Neo3     & 6 GB RAM & \makecell{Snapdragon 865 SoC:\textit{ Octa-core Kryo 585 CPU, } \\ \textit{Adreno 650 GPU and Hexagon 698 DSP}} \\ \hline
    
    Huawei nova 6 SE   & 8 GB RAM & \makecell{Kirin 810 SoC:\textit{ Octa-core CPU with two ARM} \\ \textit{Cortex-A76 and six ARM Cortex-A55 cores, } \\\textit{Mali-G52 GPU and Da Vinci NPU based DSP }} \\ \hline
    
    Google Pixel 3a  & 4 GB RAM & \makecell{Snapdragon 670 SoC:\textit{ Octa-core Kryo 360 CPU, }\\ \textit{Adreno 615 GPU and Hexagon 685 DSP}} \\ \hline
    
    Google Pixel 2 XL  & 4 GB RAM & \makecell{Snapdragon 835 SoC:\textit{ Octa-core Kryo 280 CPU, } \\ \textit{Adreno 540 GPU and Hexagon 682 DSP}} \\ \hline
    
    Google Pixel       & 4 GB RAM & \makecell{Snapdragon 821 SoC:\textit{ Quad-core CPU with two} \\ \textit{2.34 GHz Kryo cores and two 2.19GHz Kryo cores,}\\ \textit{Adreno 530 GPU and Hexagon 680 DSP}} \\ \hline

    Raspberry Pi 4B  & 4 GB RAM& Quad-Core ARM Cortex-A72 CPU\\ \hline
    
    Raspberry Pi 4B+  & 2 GB RAM& Quad-Core ARM Cortex-A72 CPU\\ \hline

     Raspberry Pi 3B+  &  1 GB RAM& Quad-Core ARM Cortex-A53 CPU\\ \hline
    
    Samsung Gear S3    & 768 MB RAM & \makecell{Exynos 7 Dual 7270 CPU } \\ \hline
    
    Raspberry Pi 1  & 256 MB RAM & ARM1176JZF-S 700 MHz CPU \\ \hline

    \makecell{Nordic Semiconductor \\nRF-52840} & 256 KB RAM& ARM Cortex-M4F CPU  \\ \hline
   
    \makecell{STMicroelectronics \\STM32F746} & 320 KB SRAM & ARM Cortex-M7 CPU \\ \hline
    
    Arduino Nano 33 BLE & 256 KB SRAM & ARM Cortex-M4 CPU \\ \hline
     
    Arduino MKR1000 & 32 KB SRAM& SAMD21 Cortex-M0+ ARM MCU \\ \hline
\end{longtable}

\begin{table}[!ht]
\caption{Overview of targeted devices and major design features of current systems.}
\centering
\rotatebox{90}{
\begin{tabular}{|c|c|c|c|} 

    \hline
    \textbf{Research} & \textbf{Device Category} & \textbf{Device} & \textbf{Design Features} \\  \hline

     LifeLearner~\cite{kwon2023lifelearner} & Low-constraint &  \makecell{NVIDIA Jetson Nano\\ Raspberry Pi 3B+} & \makecell{Continual Learning-based System,\\ Meta-CL~\cite{javed2019meta}}  \\ \hline 
    
    Sage~\cite{Sage} & Low-constraint &  \makecell{Smartphones: \textit{Samsung Galaxy S10} \\ \textit{and Samsung Note 20}} &  \makecell{Two-Stage System,\\ Micro batch~\cite{huang2019gpipe} \\and Recomputation~\cite{chen2016training}}  \\ \hline
    
    Mandheling~\cite{Mandheling} & Low-constraint & \makecell{Smartphones: \textit{Xiaomi 10, Xiaomi 11 Pro }\\  \textit{ and Redmi Note 9 Pro}}  & \makecell{Two-Stage System, \\ Digital Signal Processors}   \\ \hline
    
    Melon~\cite{Melon} & Low-constraint & \makecell{Smartphones: \textit{Samsung Note 10,  Redmi Note 8,} \\ \textit{Redmi Note 9 Pro, vivo iQOO Neo3, Meizu 16T}} & \makecell{Two-Stage System,\\ Micro batch \\and Recomputation}  \\ \hline

    zTT~~\cite{kim2021ztt} & Low-constraint & \makecell{NVIDIA Jetson TX2 \\ Smartphones: \textit{ Google Pixel 3a}}  & \makecell{Reinforcement Learning\\Based System} \\   \hline

   ElasticTrainer~\cite{huang2023elastictrainer} &  \makecell{Low-constraint \\ Middle-constraint} & \makecell{NVIDIA Jetson TX2 \\ Raspberry Pi 4B}  &  \makecell{Two-Stage System \\ Runtime
    Elastic Tensor Selection} \\ \hline

    MDLdroidLite\cite{zhang2020mdldroidlite} &  \makecell{Low-constraint \\ Middle-constraint} & \makecell{Smartphones: \textit{Huawei nova 6 SE, Google Pixel,} \\\textit{Google Pixel 2 XL and Samsung Gear S3}}  & 
\makecell{Online Continuous \\Growth-Based System} \\ \hline

    TinyTL~\cite{cai2020tinytl} & Middle-constraint & Raspberry Pi 1 & \makecell{One-Stage System,\\ Fine-tuning} \\ \hline

    MiniLearn~\cite{minilearn} & High-constraint &  Nordic Semiconductor nRF-52840  & \makecell{One-Stage System, \\Quantization~\cite{jacob2018quantization} \\ Pruning~\cite{molchanov2016pruning}} \\ \hline
    
    TTE~\cite{lin2022device} & High-constraint & STMicroelectronics STM32F746 & \makecell{Two-Stage System, \\Quantization-Aware Scaling}  \\   \hline
    
    POET~\cite{patil2022poet} & \makecell{Low-constraint  \\ Middle-constraint \\ High-constraint} & \makecell{NVIDIA Jetson TX2  \\ Raspberry Pi 4B+ \\ Arduino MKR1000 \\ Nordic Semiconductor nRF-52840}   & \makecell{Two-Stage method, \\ Mixed Integer Linear\\ Programming}  \\ 
\hline
\end{tabular}
}
\label{tab:targetdevice}
\end{table}

\subsection{One-Stage Systems}
\label{subsec:onestagesystem}
This section provides a summary and overview of two one-stage systems: MiniLearn~\cite{minilearn} and TinyTL~\cite{cai2020tinytl}. One-stage systems share a common characteristic in that they are designed to address training scenarios requiring fewer computations. Specifically, they either train simpler NN architectures, such as CNNs with fewer than ten layers, or focus on tuning only a subset of parameters within a complex NN model, such as biases in CNNs while keeping the filter parameters fixed. The main ideas behind these systems revolve around two key approaches. Firstly, there is a focus on minimizing the size of the model, as exemplified by MiniLearn's retraining of quantized models and pruning of filters~\cite{molchanov2016pruning}. Secondly, efforts are made to reduce the number of parameters requiring training. For example, TinyTL exclusively tunes biases, while MiniLearn fine-tunes only the fully connected layers. These two aspects represent common efficiency metrics in ML. It is imperative for future researchers to bear these considerations in mind, given that on-device training research inherently intersects with the realm of efficient embedded ML.

\subsubsection{Training Simple Neural Networks}
 MiniLearn re-trains Convolutional Neural Networks (CNNs) on IoT devices to empower on-device ML applications to adapt dynamically to real deployment environments, overcoming the gap between pre-collected training datasets and real-world data. The target device in this work is the nRF-52840 SoC microcontroller, which features a 32-bit ARM Cortex-M4 with FPU at 64 MHz, 256 KB of RAM, and 1 MB of Flash. MiniLearn applies quantization and dequantization techniques to enable model re-training on resource-constrained devices. The key idea is to store the weights and intermediate outputs in integer precision and dequantize them to floating-point precision during training. MiniLearn consists of three steps: (1) Dequantizing and pruning filters, (2) Training filters, and (3) Fine-tuning fully connected layers.

\fakesubpar{Dequantizing and pruning of filters} 
 First, MiniLearn performs dequantization, keeping the first layer in integer precision format to retain the input shape (which is a tensor with a specific shape) of the neural network (NN), while dequantizing the remaining layers to floating-point format. Next, during training, MiniLearn prunes less significant filters, using the L1-Norm as a static method for selecting filters. Filters with small L1 norms are considered less important in the network. Finally, MiniLearn adjusts the output layer based on the fine-tuning demand, i.e., it adjusts the number of neurons in the output layer according to the number of target output classes.

\fakesubpar{Training the filters} 
 MiniLearn preprocesses the training set and filters by collecting the output of the first layer in compressed quantized (INT8) format as the training set, which is converted to a floating-point format during training. Additionally, MiniLearn initializes models based on quantized pre-trained models and converts them to floating-point format. After pre-processing, MiniLearn performs its training operations.

\fakesubpar{Fine-tuning fully connected layers} 
 MiniLearn quantizes the pruned filters back to integer precision format and freezes them. It then performs a few extra epochs of fine-tuning on the fully connected layers to compensate for the potential loss of information caused by pruning and quantization.

\subsubsection{Tuning Part of the Neural Network}
 TinyTL aims to enable on-device training of deep CNNs on memory-constrained devices. The authors demonstrate this by training the ProxylessNAS-Mobile model~\cite{cai2018proxylessnas} on a Raspberry Pi 1, which only has 256 MB RAM. The authors argue that techniques like pruning, which reduces trainable parameters, do not fully solve the memory footprint problem in on-device training. To address this issue, TinyTL only tunes the biases of the CNN while freezing the parameters of filters to significantly reduce the memory footprint. This is because the number of bias parameters is far less than the number of filter parameters in a CNN.

 However, only tuning the biases can lead to limited generalization ability. To address this issue, the authors propose the lite residual module, which employs group convolution~\cite{NIPS2012_c399862d} and a $2\times 2$ average pooling to downsample the input feature map and reduce the number of channels. This module refines the intermediate feature maps and helps to improve the model's generalization ability. The workflow for the lite residual module is similar to the biased module. TinyTL adds the lite residual module to the CNN model to reduce the memory footprint of training while maintaining high model performance.

\subsection{Two-Stage Systems}
\label{subsec:twostagesystem}
The two-stage paradigm is popular in current systems research, with POET~\cite{patil2022poet}, Mandheling~\cite{Mandheling}, Melon~\cite{Melon}, Sage~\cite{Sage}, and TTE~\cite{lin2022device} following this paradigm. As mentioned earlier, two-stage systems have an additional preparation stage compared to one-stage systems. During the preparation stage, these systems pre-process models, generate the computing graph (DAG), and/or find an optimal or suboptimal execution plan for model training. Two-stage systems exhibit a variety of characteristics that can be categorized into three distinct types. Firstly, some systems primarily focus on addressing the memory constraints inherent in on-device training. Unlike one-stage systems, which may only fine-tune or partially train models, these systems tackle the challenge of training full-fledged NN models, such as BERT-small~\cite{devlin2018bert} and MobileNetV1~\cite{howard2017mobilenets}, where memory limitations pose the primary bottleneck. Secondly, certain systems leverage auxiliary hardware to facilitate on-device training. Recognizing that resource-constrained devices often lack powerful GPUs, these systems explore the feasibility and potential of utilizing available hardware components such as DSPs to accelerate on-device training tasks. Lastly, there are systems designed specifically to enable model training on high-constraint devices. These systems face unique challenges, including energy concerns and extremely limited memory, when training models on devices with more constrained resources.

 It is important to note that the execution plan schedules the training process by applying suitable techniques according to the characteristics of operations. However, this does not mean that the training will strictly follow one chosen schedule during runtime. The budgets of resources are dynamic during runtime, and hence, the optimal schedule is also dynamically changing. Therefore, the schedule should also be adapted correspondingly. For example, Melon generates several execution plans during the first stage and switches between them during runtime according to the dynamic runtime budgets. Melon's evaluation shows that this achieves better results than statically following one schedule.

\subsubsection{Systems Focusing on Memory Bottleneck} 
This section provides an overview of systems that target solving memory bottlenecks in on-device training, including Melon~\cite{Melon} and Sage~\cite{Sage}. While techniques explored in cloud-based systems may also be applicable to on-device training, the design of on-device training systems needs to be judicious. Melon and Sage both leverage two well-explored techniques from cloud-based systems, namely micro-batching and recomputation, to cope with memory constraints, as discussed in Section~\ref{subsec:generaltech}. These papers jointly apply these techniques to address memory limitations. However, because the micro-batch method can interfere with training results when models include batch normalization (BN) layers due to the introduction of cross-sample dependencies within a batch, additional optimization designs are introduced in on-device training systems to facilitate training. For instance, Melon employs a tensor-lifetime-aware algorithm for memory layout optimization, while Sage performs both graph- and operator-level optimizations for the computation graph (DAG).
 \\
\fakepar{Melon}
Melon is designed based on the observation that constrained memory hinders training performance. Specifically, large enough batch sizes and batch normalization layers are two key factors for guaranteeing the convergence accuracy of large NNs. However, resource-constrained devices cannot support large batch sizes during training, as this leads to large peak memory usage. When training a model with batch normalization layers, the mean and deviation of small-batch samples are not representative enough of the distribution of the whole dataset. This slows down the training process and reduces the model's generalization ability. Melon aims to solve these two problems. The authors present a memory-calibrated progressive recomputation method based on the tensor-lifetime-aware memory pool. The method takes the whole operator DAG as input, and to estimate the benefit of recomputing each tensor, the authors define a metric called Triangle Per Second (TPS) as shown in Equation~\ref{eq:tps}. 

\begin{equation}
    TPS = \frac{TensorSize \times FreedLifetime}{ResomputationTime}
\label{eq:tps}
\end{equation}
 where $TensorSize$ refers to the size of activations, $FreedLifetime$ denotes the lifetime span between discarding and recomputing, and $ResomputationTime$ indicates the time it costs to recompute the discarded activation.

Melon integrates micro-batch~\cite{huang2019gpipe} and recomputation~\cite{chen2016training} to reduce memory consumption. The main idea is to minimize the recomputation overhead by scheduling when and which tensors (intermediate output of hidden layers) should be discarded or recomputed. Melon employs a tensor-lifetime-aware algorithm for memory layout optimization since the performance of recomputation highly depends on memory management.  Melon defines the tensors into two categories and manages them respectively. Firstly, for tensors with long lifetimes, such as a tensor produced at the forward pass and released at the backward pass, Melon follows a First Produced Last Release order. For others, it follows a greedy way to produce and release them. Overall, the key idea is to place those long-lifetime tensors beneath short-lifetime ones to consolidate the overall memory layout. When performing recomputation, Melon will continuously discard tensors with maximal TPS and calibrate the memory pool until the pool size is smaller than the budget.

\fakesubpar{First stage} 
 Melon denotes its first stage as the decision stage. At this stage, it generates the optimal execution plan based on the given budget. Specifically, it obtains runtime information via an execution profiler that contains NN operators and intermediate tensors. The tensor information consists of data flow dependencies, the computation time of each operator, as well as the size and lifetime of each tensor. Based on the profiler, Melon generates an execution plan that determines (1) where each tensor is placed in a large memory pool, (2) which operators need to be recomputed, and (3) how to split the batch. Melon generates multiple plans for different budgets but does so only once.

\fakesubpar{Second stage} 
 In the execution stage, Melon performs the training based on a proper plan. Notably, Melon can switch to a more proper plan when the memory budget changes. The switching strategy is a "lazy strategy", i.e., switching to a new plan when the current training batch ends.

\fakepar{Sage}
 Sage aims to enable SOTA  DNN training on smartphones. Specifically, the authors implemented Sage to train ResNet-50~\cite{he2016deep}, DenseNet-121~\cite{huang2017densely}, MobileNetV2~\cite{sandler2018mobilenetv2}, and BERT-small~\cite{devlin2018bert} on smartphones with different SoCs, including Samsung Galaxy S10 and Note 20. The authors also demonstrate that the scarce and dynamic memory of mobile devices is the major bottleneck of on-device training, with memory usage during training being 5 to 100 times higher than during inference. To address this issue, Sage employs dynamic gradient checkpointing~\cite{kirisame2020dynamic} and dynamic gradient accumulation to dynamically adjust the memory usage to the available system memory budget. Dynamic gradient checkpointing shares the same idea with materialization, dropping intermediate results and recomputing them when needed. For dynamic gradient accumulation, the authors use a traditional micro-batch technique~\cite{stein2021latency} that can dynamically adjust the size of the micro-batch according to the available memory during runtime to reduce peak memory usage.

\fakesubpar{First Stage} 
 In its first stage, Sage constructs a computation graph (DAG) with automatic differentiation. Concretely, a DNN node is expressed by differentiable operations and computable operations abstraction. In this way, the DAG decouples the evaluation and differentiation processes of DNN computation graphs. Then, Sage performs both graph- and operator-level optimizations for the DAG.

\fakesubpar{Second stage} 
 This stage is also called the graph execution stage. Sage employs a hybrid strategy to combine gradient checkpointing and gradient accumulation during runtime adaptively to execute the DAG. 

\subsubsection{Systems Leveraging Auxiliary Hardware}
\label{subsec:mandheling}
Hardware acceleration is a common approach as a general training speedup technique. However, resource-constrained devices typically lack accelerators dedicated to on-device training, which are prevalent in many specialized servers. As discussed in Section~\ref{subsec:scope}, some devices, such as smartphones, are equipped with hardware accelerators, e.g., mobile GPUs. Nevertheless, these accelerators typically do not possess dedicated peripherals or resources, such as memory and power. Instead, they share many of the resources with other components consisting the system. This means that the excessive use of resources at the accelerators can lead to overall system inefficiency and potentially lead to degradation of the user experience~\cite{shingari2015characterization, pramanik2019power}.

As an alternative, on-device training systems can leverage other auxiliary hardware to accelerate their training tasks, for example, Mandheling~\cite{Mandheling} enables on-device training with DSP offloading. By utilizing DSPs or similar hardware, on-device training systems can overcome the limitations imposed by the absence of GPUs and improve the efficiency of model training on resource-constrained devices.

 \fakepar{Mandheling} is designed to train DNNs on modern smartphones. The author's starting point is that DSPs are particularly suitable for integer operations, and DSPs are ubiquitously available on modern smartphones. Therefore, the authors leverage CPU-DSP co-scheduling with mixed-precision training algorithms to enable on-device training.

\fakesubpar{First stage} 
 Denoted as the preparing stage, here, Mandheling initiates the model transformation. The target models could either be pre-trained or randomly initialized based on different frameworks, such as TensorFlow and PyTorch. Given a model, Mandheling translates it into FlatBuffer-format, a format with quantization used in TensorFlow Lite. In the execution stage, Mandheling generates CPU and DSP compute subgraphs, then performs compute-subgraph execution on devices. 

\fakesubpar{Second stage} 
 In the execution stage, Mandheling fully leverages DSPs for efficient model training by using four key techniques: (i) CPU-DSP co-scheduling, (ii) self-adaptive rescaling, (iii) batch splitting, and finally, (iv) DSP-computes subgraph reuse.

 CPU-DSP co-scheduling reduces the overhead of CPU-DSP context switching by avoiding the offloading of DSP-unfriendly operators to the DSP. Examples of such operators include normalization, dynamic rescaling, and compute-graph preparation. Normalization is considered DSP-unfriendly due to its irregular memory access.

 Self-adaptive rescaling refers to periodically performing rescaling instead of performing it every batch. Rescaling involves adjusting the scale factor, which is a trainable parameter used to scale the output of each layer. Dynamic rescaling adjusts the scale factor per batch and guarantees convergence accuracy but adds at least two times the latency compared to static rescaling, which uses a fixed scale factor. In their experiments, the authors observed that the scale factor jumps between 10 and 11, and the frequency of these changes is low, occurring once every 10 to 60 batches. Based on this observation, the authors propose to rescale periodically instead of after each batch.

 The batch-splitting phase splits batches based on their workload. Specifically, Mandheling first identifies abnormal batches with noticeably higher workloads, then splits these batches into multiple micro-batches to execute them individually.  
 DSP-compute subgraph reuse is the fourth technique used in Mandheling. The DSP-compute subgraph comprises operators with inputs, outputs, and parameters. In the current training method, a new compute subgraph is prepared for each training batch. However, the authors find that models are rarely modified during training. Therefore, reusing the compute subgraph eliminates the preparation overhead. Additionally, Mandheling releases the most recently used (MRU) memory when the memory runs out. This is because the allocation/deallocation of memory for subgraph reuse always follows the execution order of DNNs, which means that the MRU memory region has the longest reuse distance.

\subsubsection{Systems Designed for High-Constrained Devices} 
This section provides an overview of systems designed to enable model training on high-constrained devices, such as Private Optimal Energy Training (POET)~\cite{patil2022poet} and Tiny Training Engine (TTE)~\cite{lin2022device}. These systems focus on optimizing the training process and applying general optimization techniques to overcome the constraints of resource-intensive tasks on such devices. POET approaches the challenge by abstracting training memory and computation costs into Mixed Integer Linear Programming (MILP) problems. By solving these MILP problems, POET generates a training schedule optimized for efficiency. Additionally, POET applies techniques like paging and gradient checkpointing to further optimize the training process. On the other hand, TTE takes a different approach by separating auto-differentiation from runtime and moving it to compile time. By doing so, TTE aims to streamline the training process and improve efficiency. Additionally, TTE leverages pruning techniques to accelerate training even further, making it suitable for high-constrained devices where computational resources are limited.

\fakepar{POET}
 POET aims to enable the training of SOTA neural networks on memory-scarce and battery-operated edge devices. The authors devised POET to allow for on-device training of DNNs, such as ResNet-18~\cite{he2016deep} and BERT~\cite{devlin2018bert}. The target devices include the ARM Cortex M0 class MKR1000, ARM Cortex M4F class nrf52840, A72 class Raspberry Pi 4B+, and Nvidia Jetson TX2.

 \fakesubpar{First stage} Firstly, given an NN model, POET profiles the memory and computation costs of its operators. Operators refer to the computational operations of the ML model, such as non-linear functions and convolutions. Based on the resource cost of operators, POET selects a suitable technology to optimize them during training. Secondly, taking into account hardware constraints, POET generates a Mixed Integer Linear Programming (MILP) problem to search for the optimal schedule of paging~\cite{tanenbaum2009modern} and rematerialization~\cite{chen2016training}. Finally, POET sends the schedule to the target edge devices to carry out memory-efficient ML training. The objective function of the MILP is denoted as follows:

\begin{equation}
    min \sum_{T}[R \Phi_{compute}+M_{in} \Phi{pagein} + M_{out}\Phi_{pageout}]_T
\end{equation}

 where $\Phi_{compute}$, $\Phi_{pagein}$, and $\Phi_{pageout}$ denote the energy consumption of each computing node, page in, and page out operations, respectively. $R$ refers to recomputing operations, $M_{in}$ represents paging a tensor from secondary storage to RAM, and $M_{out}$ refers to the opposite operation.

\fakesubpar{Second stage}  POET performs model training by following the paging and rematerialization schedule generated in the first stage.

\fakepar{TTE}
 The authors present the TTE to tackle the two main challenges of on-device training, namely: (1) the difficulty in optimizing quantized models. The deployed model is quantized, which leads to lower convergence accuracy than the unquantized model due to low precision and lack of batch normalization layers. (2) Limited hardware resources of tiny devices. The memory usage of full back-propagation can easily exceed the memory of microcontrollers. The authors conduct experiments using three popular TinyML models: MobileNetV2, ProxylessNAS, and MCUNet~\cite{lin2020mcunet}. The experiments are performed on a microcontroller STM32F746 that features 320KB SRAM and 1MB Flash using a single batch size. The authors verify that the quantization process distorts the gradient update, causing the lower convergence accuracy problem of quantified models. To address this problem, they propose quantization-aware scaling (QAS). The key idea of QAS is to multiply the square of scaling factors corresponding to precision with intermittent output to relieve the disorder caused by quantization. Then, TTE combines QAS and sparse update techniques to overcome the second challenge.

 \fakesubpar{First stage}
 In the first stage, the TTE engine works in three steps. Firstly, TTE moves the auto-differentiation to the compile time from the runtime and generates a static backward computing graph. Secondly,  TTE prunes away the gradient nodes with the sparse layer update method. Thirdly, TTE reorders the operators to immediately apply the gradient update to a specific tensor before back-propagating to earlier layers so that the memory occupied by the gradient tensors can be released as soon as possible. 
 
\fakesubpar{Second stage} In the second stage, TTE executes the computing graph from the first stage.

\subsubsection{Systems Focusing on Training Speedup} 
 On-device training can be time-consuming due to constrained hardware resources. Hence, training speedup is essential for on-device training. To speed up the on-device training process, previous research focuses on three main approaches. The first approach offloads the entire DNN training operation to the cloud and then only trains a selected portion of DNN on the device~\cite{donahue1deep, sharif2014cnn}, by performing transfer learning. However, this approach reduces the learning capability of models, which leads to a drop in accuracy~\cite{cai2020tinytl}. The second approach starts with a tiny structure and gradually grows the DNN during training until satisfactory accuracy is achieved. This method is called continuous growth~\cite{irsoy2019continuously}. Continuous growth has been applied in MDLdroidLite~\cite{zhang2020mdldroidlite}, which we discuss in the next section. However, a continuous growth-based system must train newly added NN layers from scratch, requiring over 30\% extra training epochs~\cite{zhang2020mdldroidlite}. The third approach adaptively adjusts the trainable portion of a DNN. This approach gradually removes less important layers on the fly during training~\cite{he2017channel, molchanov2019importance}. However, this selection of layers is an unrecoverable process, which means that the removed pruned portions can never be selected again even if they may have more importance later. Therefore, this approach cannot flexibly adapt to changing requirements.  

\fakepar{ElasticTrainer}
ElasticTrainer~\cite{huang2023elastictrainer} presents an elastic runtime adaptation approach for speedup of on-device training, where every DNN substructure can be freely removed from or added to the trainable DNN portion at any time in training. To achieve elasticity, the key is to select the optimal trainable NN portion at runtime, which meets the desired training speedup with the minimum accuracy loss. The authors design a two-stage system to achieve this target.

\fakesubpar{First stage} In the first stage, ElasticTrainer exploits a tensor timing profiler to profile the estimated training time for the selected DNN sub-architecture. Concretely, the authors first convert the original DNN structure into a tensor-level computing graph, which retains the execution order of all tensor operations during training. Then, they use a standard DNN profiler~\footnote{https://www.tensorflow.org/guide/profiler} to measure the execution time of each operation, such as matrix multiplication and convolution. Finally, the training time of each tensor is computed according to the timings of related operations.

\fakesubpar{Second stage} In the second stage, ElasticTrainer considers both accuracy and training time in the selection of the optimal trainable DNN portion. However, it is practically difficult to determine whether the required accuracy could be achieved within reasonable training time in advance. Therefore, ElasticTrainer uses the training loss reduction at the desired training time as the selection metric instead. ElasticTrainer defines the selection problem as a constrained optimization problem as follows:

\begin{equation}
\max \Delta_{\text{loss}}(\mathcal{M}) \quad \text{s.t.} \quad T_{\text{selective}}(\mathcal{M}) \leq \rho T_{\text{full}},
\end{equation}
where $\mathcal{M}$ refers to a binary mask for tensor selection, with its $j$-$th$ element indicating if the $j$-$th$ tensor is selected. For example, if $\mathcal{M} = [0, 1, 1, 0]$, the second and third tensors will be trainable tensors, and others will not be trained. $T_{\text{selective}}$ denotes the total accumulated estimated training time of selected tensors. The estimated training time of each tensor is computed by the tensor timing profiler in the first stage. $T_{\text{full}}$ refers to the full training time (the time required to train the full DNN). $\rho$ depicts the expected training time reduction ratio, with $\rho = 50\%$ referring that the training time should be reduced to half of the full training time.

\subsection{Continual Learning-Based Systems}
\label{subsec:MDLdroidLite}
Continual learning enables DNN models to learn new tasks over longer durations. These models retain and use the knowledge learned from previous tasks to learn new tasks more efficiently and effectively~\cite{silver2013lifelong, parisi2019continual}. However, they generally suffer from the catastrophic forgetting (CF) issue~\cite {mccloskey1989catastrophic}, which means the model severely forgets previously learned knowledge while learning new data. 
To alleviate CF, researchers propose three common approaches~\cite{parisi2019continual}: regularization approaches, dynamic architectures, and memory replay. Regularization approaches impose constraints on updating the weights of neural networks to alleviate CF. This method generally penalizes differences between the weights for the old and the new tasks to slow down the update of task-relevant weights. The dynamic architecture approach dynamically adapts the model's architecture in response to new tasks, such as re-training the model with an increased number of neurons or network layers. MDLdroidLite~\cite{zhang2020mdldroidlite} uses the dynamic architectures approach in its on-device training process. The memory replay approach leverages a dual-memory learning mechanism to cope with CF, which is inspired by the complementary learning systems theory~\cite{mcclelland1995there, kumaran2016learning}. In the dual-memory learning mechanism, the model has two types of weights: fast change weight for temporary knowledge and slow change weight, which stores long-term knowledge~\cite{hinton1987using}. LifeLearner~\cite{kwon2023lifelearner} and Miro~\cite{ma2023cost} both apply the memory replay approach. However, as mentioned before, the mismatch between hardware resource availability and demand is one of the key challenges of on-device training. Continual lifelong learning-based on-device training systems also optimize the continual learning approaches for resource efficiency. 

\fakepar{MDLdroidLite} MDLdroidLite uses the dynamic architectures approach for on-device training. Concretely, MDLdroidLite starts with a tiny structure, which is only 1\%-10\% of the full model size, and gradually grows the DNN during training until achieving satisfactory accuracy. The method is called continuous growth. The inspiration is that traditional DNN models often have large redundant parameters that can be reduced through pruning techniques, as shown in previous studies~\cite{frankle2018lottery}. This continuous growth approach has been explored in previous works that apply knowledge transfer to fast-tune newly grown neurons or layers using techniques such as net2net~\cite{dai2019nest, du2019cgap, irsoy2019continuously}. However, current methods may still generate models with redundant parameters that do not fit mobile devices well and exhibit slow convergence rates that cannot guarantee convergence on resource-constrained devices. 

To address these challenges, the authors introduce a release-and-inhibit (RIC) control and RIC-adaption pipeline. The RIC control manages the layer-level growth of DNN models to enable independent layer growth and avoid redundant parameter growth. The RIC-adaption pipeline addresses the slow convergence rate issue by adapting the RIC control during training. 

\fakesubpar{RIC control} 
 The reward function used in the RIC control of MDLdroidLite is defined as the difference in accuracy achieved by growing a layer and the accuracy achieved without growing the layer. More formally, the reward function is defined as follows:

\begin{equation}
    Reward = \frac{G(\cdot)}{C(\cdot)}
\end{equation}
 
 where $G(\cdot)$ and $C(\cdot)$ refer to the state-value and state-cost function, respectively. These two functions measure the difference in value and cost between states before and after one growth action. The state-value function measures the gains of growth and the state-value function measures the extra cost from growth, that defines as $ C(\cdot) = (1-\beta)Flops(\cdot) + \beta Size(\cdot)$, where $Flops(\cdot)$ and $Size(\cdot)$ denote FLOPs of the grown layers and the size of the grown layers respectively. $\beta$ refers to a normalization coefficient.

\fakesubpar{RIC adaption pipeline} 
 To overcome the low convergence rate caused by parameter growth, the authors propose a three-step adaptation pipeline that safely adapts newly added parameters without sacrificing accuracy in each growth step. Firstly, they select neurons or channels with small variance using a cosine similarity function to add to the NN. Secondly, they preserve loss reduction by applying a safe parameter scale function. Specifically, they initialize the parameters using the Kaiming method~\cite{he2015delving}, which is based on the distribution of parameters in the full-sized model. Lastly, they map the layer with newly added neurons to subsequent layers to maintain the relationship between layers. This three-step adaptation pipeline ensures that the newly added parameters are adapted safely and effectively, enabling faster convergence and higher accuracy. Overall, MDLdroidLite enables the on-device training of SOTA  DNNs from scratch. 

\fakepar{LifeLearner} LifeLearner leverages meta-continual learning (Meta-CL) and the memory replay approach for its on-device training operations. Meta-CL is a specialized branch of continual learning that incorporates meta-learning techniques to enhance adaptability and mitigate forgetting across sequential tasks~\cite{javed2019meta, beaulieu2020learning, riemer2018learning}. Meta-learning helps create a model that can quickly adapt to new tasks with minimal updates, which consists of two phases: meta-training and meta-testing. The meta-training phase seeks to find an optimal initialization for weights of the DNN model, which is generally performed on offline servers. The meta-testing phase further tunes the initialized DNN model with a few new data samples. This phase could run on embedded computing platforms. To alleviate the forgetting problem, prior works in Meta-CL~\cite{beaulieu2020learning, javed2019meta, lee2021few} split the DNN architecture into a feature extractor and a classifier, and adapt the idea of memory replay at an architecture level. During the meta-training phase, prior works in Meta-CL update the feature extractor (slow change weights) in an outer loop with random samples from learned classes to store long-term knowledge, and the classifier (fast change weights) in the inner loop (fast weights) to learn new temporary knowledge. However, after deployment, prior works in this domain rely on the inner loop (updating the classifier) in the meta-testing phase for learning, which leads to limited generalization capability~\cite{kwon2023lifelearner}. Furthermore, those methods incur high computation and memory costs. 

To address these challenges, the authors of LifeLearner first design a deployment-time training optimization mechanism for utilizing memory replay-based Meta-CL to resolve the accuracy degradation problem, referred to as the Co-utilization of Meta-Learning and Rehearsal Strategy. Secondly, they present an optimization method that provides a computation-efficient replay strategy to minimize the computation overhead along with a compression module for the compression and storage of replay samples to induce memory usage, termed as CL-tailored Algorithm/Software Co-Design.

\fakesubpar{Co-utilization of Meta-Learning and
Rehearsal Strategy} The authors introduce a replay buffer for storing samples of classes already learned, which are mixed with samples from new classes. This buffer prevents the catastrophic forgetting issue of continual learning. In addition, they design a deployment-time training optimization mechanism. After deployment, the system performs inner-loop training with samples of new classes to learn new temporary knowledge and also outer-loop training with samples from learned classes for retaining long-term knowledge. This mechanism brings extra costs both in computation and memory. To alleviate the overhead cost, the authors present a compression module as described below.

\fakesubpar{CL-tailored Algorithm/Software Co-Design} LifeLearner exploits a replay strategy and a compression module to minimize the overhead cost from the deployment-time training optimization mechanism. Concretely, the replay strategy is to store intermediate activations instead of raw data samples. LifeLearner stores the output of the last layer of the model’s feature extractor as replay data samples for inner-loop updating. During inner-loop updating, it freezes the feature extractor and only performs training on the classifier. In addition, on the memory front, the authors observe that the activations are sparse due to using the ReLU non-linearity activation function, i.e., more than 90\% of the activation values of the hidden layer are zero. The sparsity facilitates the compression and subsequent efficient storage of activations on-device. 

Therefore, the authors design a compression module for replay samples. The compression module consists of two stages: sparse bitmap compression and product quantization. The sparse bitmap compression stage filters out the majority of zero values in activations. Specifically, for compression,  it firstly creates a bitmap with the same dimensions as the activations, then sets a bit to 1 for non-zero values’ indices and 0 for the remainders, then stores the non-zero values in a vector with 32-bit floats format and the indices bitmap in bitmap format. For decompression it firstly traverses the indices bitmap and the non-zero values vector, then reconstructs the activations by using either the saved non-zero value or zero according to the indice is 1 or 0. In the second stage, the authors leverage production quantization~\cite{jegou2010product} to further compress the vector stores non-zero values in the first stage. 

\fakepar{Miro} 
As previously mentioned, the authors of Miro only evaluate their system using NVIDIA Jetson platforms, which is outside the scope of this survey. However, we believe that Miro holds potential for further optimization. Future research could explore its feasibility on more resource-constrained devices, such as Raspberry Pi. Therefore, we also provide a summary of the Miro system. Concretely, Miro enables on-device training by leveraging and optimizing hierarchical episodic memory (HEM)~\cite{lee2022carm}, an optimized variant of episodic memory (EM)~\cite{lopez2017gradient}. EM is a memory replay method that alleviates the CF issue. With EM, the system stores old data samples in the memory and replays them together with new data samples during continuous learning. However, EM incurs large memory costs that limit its applications in hardware-constrained devices. HEM is an optimized variant of EM that leverages hierarchical memory to reduce memory usage. Concretely, HEM stores a small set of old data samples in memory with fast access and a large number of old data samples in storage with slow access. During training, HEM fetches old samples in memory to train the model and performs an online data swapping to swap old data samples between memory and storage. With this strategy, HEM balances efficiency and accuracy. 

\subsection{Reinforcement Learning-Based Systems}
\label{subsec:ztt}
Indeed, reinforcement learning (RL) presents a promising approach for on-device training. In RL, an agent learns to make decisions by interacting with an environment to maximize some notion of cumulative reward~\cite{sutton2018reinforcement}. This iterative process of adjusting the agent's policy based on feedback from the environment aligns well with the concept of on-device training. By leveraging RL techniques, on-device training systems dynamically adapt their model training strategies based on real-time feedback from the device's environment. This allows for a flexible and adaptive learning process, particularly suitable for resource-constrained devices where traditional training methods may not be feasible due to limited computational resources or memory constraints. Overall, RL represents a powerful paradigm for on-device training, offering the potential for efficient and adaptive model training tailored to the specific constraints and requirements of resource-constrained devices.

\fakepar{zTT} zTT~\cite{kim2021ztt} presents a novel approach to DVFS for mobile devices, which employs Deep Q-Learning (DQN)~\cite{mnih2013playing, gu2016continuous, wang2016dueling} to overcome the shortcomings of traditional DVFS techniques. Previous DVFS methods for mobile devices, including algorithmic-based approaches, graphics-based approaches, dynamic techniques, and thermal-aware energy management, have encountered various difficulties: Firstly, they are not capable of adjusting power distribution between CPU and GPU based on application Quality of Experience (QoE) because they are application-agnostic. Secondly, most DVFS methods lead to overheating issues in mobile devices due to a lack of cooling methods. Thirdly, supervised learning-based DVFS methods are limited in their performance because mobile devices operate in complex and changing external environments. ZTT tackles these limitations by leveraging DQN to learn and adapt power distribution based on actual performance, while also managing heat generation and adjusting to changing external environments.

The authors' proposed system is based on three key observations. Firstly, the performance of mobile device applications relies heavily on the CPU and GPU and can be optimized by controlling their clock frequencies using DVFS methods that consider the specific performance characteristics of each application, such as CPU- or GPU-intensive tasks. Secondly, the temperature of the mobile processor rises with increased heat generation due to the thermal coupling between the CPU and GPU in the confined space of mobile devices. Finally, mobile devices are susceptible to environmental changes, such as ambient temperature and whether the device is in a protective case.

\fakesubpar{System Design}
The authors designed and implemented zTT using DQN to address the aforementioned issues. Their aim is to obtain the optimal policy that controls the CPU/GPU frequency to achieve the best possible performance with the least power consumption. The definition of performance may vary depending on the application type. In this study, the authors focus on video and gaming applications as the target applications and employ frame rate as the performance metric. The design of the Deep RL-based method is outlined below.

\noindent{State:} 
zTT defines a state using seven parameters: $\{f_C(t), f_G(t), T_C(t), T_G(t), P_C(t), P_G(t), x(t)\}$, where $f_C(t)$ and $f_G(t)$ refer to the configured frequency of CPU and GPU at time step $t$, $T_C(t)$ and $T_G(t)$ denote the temperature of CPU and GPU at time step $t$, $P_C(t)$ and $P_G(t)$ represent the power consumption of CPU and GPU at time step $t$, and $x(t)$ refers to the frame rate at time step $t$.

\noindent{Action:} 
The action set of zTT consists of three actions: exploration, exploitation, and cool-down. Actions adjust the clock frequency of the CPU and GPU. Concretely, zTT follows the $\epsilon$-greedy fashion. Within the temperature threshold, zTT respectively takes exploration and exploitation action with probability $\epsilon$ and $1-\epsilon$. When the temperature gets close to the threshold, zTT does a cool-down action that randomly selects clock frequencies lower than the current ones.

\noindent{Rewards:} 
The reward function is defined as Equation~\ref{eq:reward}, where $U(t)$ refers to the performance (frame rate) at time step $t$, $P(t)$ represents power consumption at time step $t$ and $\beta$ denotes a trade-off weight between performance and power consumption. In addition, $W(t)$ is a compensation function that gives a positive reward to actions that result in temperature under the threshold and a negative reward to actions that lead to temperature exceeding the threshold.

\begin{equation}
    R = U(t) + \frac{\beta}{P(t)} + W(t)
\label{eq:reward}
\end{equation}

\noindent{Neural Networks:} 
 zTT is based on transfer learning. The approach applies a pre-trained fully connected neural network model as the initial Q-function.

\section{Review from the Machine Learning Perspective}
\label{Sec:ml}
This section analyzes current systems from the ML perspective. The analysis focuses on two aspects: the categories of NNs evaluated in the selected papers and the training strategy.

\subsection{Evaluated Neural Networks}
 Table~\ref{tab:targetNN} shows neural networks used to evaluate on-device training systems, which consist of four categories: convolutional neural networks (CNN), transformer~\cite{vaswani2017attention}-based NN, unsupervised learning NN, and neural architecture search (NAS)-based NN. Generally, these neural networks cover the most widely used NN in various fields. Concretely, CNNs include VGG~\cite{simonyan2014very}, ResNet~\cite{he2016deep}, InceptionV3~\cite{szegedy2016rethinking}, DenseNet~\cite{huang2017densely}, MobileNet~\cite{howard2017mobilenets}, SqueezeNet~\cite{iandola2016squeezenet}, and LeNet~\cite{lecun1998gradient}. The transformer-based NNs include BERT~\cite{devlin2018bert}, BERT-small~\cite{devlin2018bert}, and Vision Transformer (ViT)~\cite{dosovitskiy2020image}.
 Evaluated unsupervised learning NN is Autoencoder~\cite{hinton1993autoencoders}, a kind of unsupervised learning technique for representation learning. It is widely used in knowledge distillation~\cite{gou2021knowledge}. In addition, the NAS-based NNs include ProxylessNAS~\cite{cai2018proxylessnas}, ProxylessNAS-Mobile, and MCUNet~\cite{lin2020mcunet}. NAS is widely used in the automatic design of NNs and outperforms hand-designed architectures in many cases~\cite{zoph2016neural, zoph2018learning}. In particular, POET~\cite{patil2022poet}, Mandheling~\cite{Mandheling}, Sage~\cite{Sage}, Melon~\cite{Melon}, TTE~\cite{lin2022device}, and MDLdroidLite~\cite{zhang2020mdldroidlite} are all evaluated by training various CNNs. In addition, MiniLearn~\cite{minilearn} is designed for training CNNs on devices. ElasticTrainer~\cite{huang2023elastictrainer} is evaluated with CNNs models and ViT. In addition, for a fair comparison, LifeLearner~\cite{kwon2023lifelearner} employs CNNs for evaluation, which are used in its prior relevant works. Apart from CNNs, BERT is another popular model for on-device training works, such as POET and Sage, which is widely used and significantly successful in natural language processing tasks. Furthermore, the authors of TinyTL~\cite{cai2020tinytl} and TTE~\cite{lin2022device} also evaluated their system on ProxylessNAS. 

 \begin{table*}
  \caption{Overview of evaluated Neural Networks by current systems research.}
  \label{tab:targetNN}
  \begin{tabular}{|c|l|}
    \hline
   \textbf{Systems} & \textbf{ Neural Networks}\\
    \hline
    POET & VGG16~\cite{simonyan2014very}, ResNet-18~\cite{he2016deep} and BERT~\cite{devlin2018bert}\\ \hline
    
    Mandheling & VGG-11/16/19~\cite{simonyan2014very}, ResNet-18/34~\cite{he2016deep}, LeNet~\cite{lecun1998gradient} and InceptionV3~\cite{szegedy2016rethinking} \\ \hline
    
    Sage & ResNet-50~\cite{he2016deep}, DenseNet-121~\cite{huang2017densely}, MobileNetV2~\cite{sandler2018mobilenetv2} and BERT-small~\cite{devlin2018bert}  \\ \hline
    
    Melon &   MobileNetV1~\cite{howard2017mobilenets}, MobileNetV2, SqueezeNet~\cite{iandola2016squeezenet} and ResNet-50\\ \hline

    TinyTL & ResNet-50 and  MobileNetV2\\  \hline

    MiniLearn & \makecell[tl]{CNNs with three to four convolutional layers and two to three fully \\connected layers}\\  \hline

    TTE & MobileNetV2, ProxylessNAS~\cite{cai2018proxylessnas} and MCUNet~\cite{lin2020mcunet}\\ \hline

    zTT & Deep Q Network~\cite{mnih2015human}\\ \hline

    MDLdroidLite & LeNet, MobileNet and VGG \\ \hline

    LifeLearner & CNNs with fully-connected layers at the end \\ \hline

    ElasticTrainer & ResNet-50, VGG16, MobileNetV2 and Vision Transformer~\cite{dosovitskiy2020image}\\ 
    \hline
  \end{tabular}
\end{table*}

\subsection{Deep Neural Network Training Strategies}

\subsubsection{Transfer Learning}
 In terms of training strategies, all systems, except for Mandheling, utilize transfer learning. The authors of Mandheling perform both transfer learning and training from scratch, which involves initializing the model with random numbers. The authors conclude that transfer learning is more suitable for resource-constrained devices as it converges in fewer iterations, resulting in lower resource usage.

\begin{figure}[!ht]
    \centering
    \includegraphics[width=0.6\linewidth]{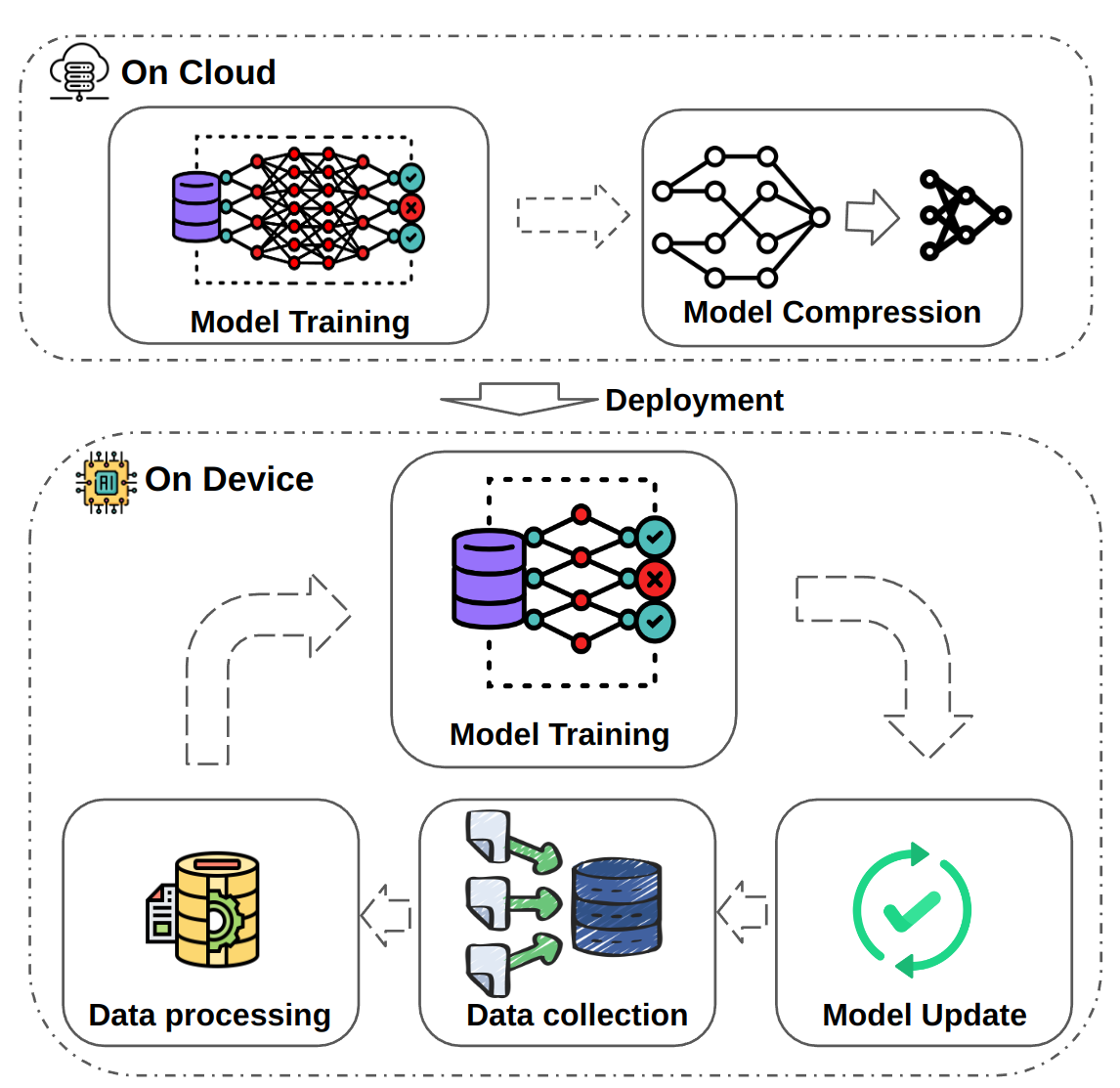}
    \caption{A generic workflow of on-device training.}
    \Description{A generic workflow of on-device training.}
    \label{fig:workflow}
\end{figure}

 The workflow of transfer learning involves first pre-training models on cloud/servers, then deploying the model, and finally, updating the model through training on the device. Training models from scratch is too resource-intensive and energy-consuming for devices, even with SOTA optimization methods~\cite{Mandheling}. However, it is unnecessary to train models from scratch as the ML community has presented numerous NNs, datasets, and pre-trained models. Depending on the fact at hand, system researchers can select a suitable NN and train a model in the cloud on a large and general dataset, then deploy the model to devices and fine-tune the model on a new and small dataset. This workflow is more efficient and practical than deploying a randomly initialized model to the device and training from scratch. Figure~\ref{fig:workflow} presents a generic workflow of on-device training.

\subsubsection{Reinforcement Learning}
 zTT is implementing the on-device training application using DQN, an ML-based variant of Q-learning~\cite{watkins1989learning} that employs an NN model to approximate the Q-function. Q-learning is a value-based reinforcement learning algorithm~\cite{kaelbling1996reinforcement}, in which the agent takes actions leading to optimal accumulated Q-values. The paradigm of reinforcement learning (RL) differs from that of standard supervised learning. Supervised learning requires a dataset with ground truths, and data samples are generally assumed to be independent and identically distributed (i.i.d.). However, in RL, the learning process depends on the interaction between the agent and the environment without ground truths. Furthermore, data samples are not i.i.d. because previous outputs influence future inputs. Specifically, during the RL learning process, the agent takes an Action based on the State of the environment and receives a Reward from the environment after taking the Action; the Action, in turn, affects the State of the environment, and the agent adjusts the policy according to the Reward. The learning process does not necessitate ground truth (or labels), which is a core advantage of RL.

 In zTT, the authors exploit a pre-trained, fully connected NN model to estimate the Q-function. The training process follows three steps. Firstly, the DQN agent takes action, calculates the reward, and observes the resulting next state. Secondly, it collects samples and puts them into the replay memory until termination. Samples consist of state, action, reward, and next state. Thirdly, random batch sampling from the replay memory and using these samples to train the NN model.

\section{Training optimization strategies}
\label{Sec:optimaztiontechs}
To reduce memory consumption, the surveyed works apply or combine different optimization strategies, which can be categorized as (i) memory optimization, (ii) model optimization, and (iii) additional-hardware assistance.

\subsection{Memory Optimization}
 All of the works mentioned above, with the exception of TinyTL, enable on-device training by utilizing various memory optimization techniques. During model training, peak memory usage is significantly higher than during inference, ranging from four to 103 times larger~\cite{qu2022p, Sage, lin2022device}, primarily due to a large number of activations, including those of the hidden layers. To avoid out-of-memory issues and improve training throughput, memory optimization techniques are essential for on-device training. Current works mainly employ recomputation, paging, and micro-batch techniques for this purpose.

\fakepar{Recomputation} 
 Recomputation is a memory optimization technique that involves deleting intermediate activations to free up memory and then recomputing them when needed, such as the output of hidden layers and the results of a non-linear function. POET, Melon, and Sage are examples of works that employ recomputation techniques to reduce their peak memory usage during on-device training.

\fakepar{Paging} 
 Paging is a memory management technique widely used in various operating systems. It divides physical memory into fixed-size pages and logical memory of a program into equally sized "page frames". Each page frame is mapped to a physical page. During the program execution, the OS retrieves the needed page frame from secondary storage if it is not in physical memory. Meanwhile, the OS pages out the unneeded page frame to the secondary storage from memory in order to release memory. POET utilizes paging as a memory management technique to reduce peak memory usage by paging out the unneeded intermediate activations into the secondary storage during on-device training.

\fakepar{Micro-batch} 
 Micro-batch is a training optimization technique that involves using smaller batch sizes during training to reduce peak memory usage. However, using a smaller batch size may negatively impact the performance of the model with batch normalization layers. As a result, there is a trade-off between feasibility and accuracy when using this technique. Mandheling, Sage, and Melon are examples of works that utilize micro-batch techniques in their systems for on-device training optimization.

\subsection{Model Optimization}
 In addition to the memory optimization techniques described above, on-device training systems also leverage model-side optimization techniques, with a target to simplify the model to reduce memory consumption and the number of floating number operations (FLOPs). These techniques include quantization, dequantization, mixed-precision, sparse updates, continuous growth (CG), and adjusting the architecture of neural networks, such as replacing or inserting new layers into the network. These techniques are used to further optimize the on-device training process and improve the efficiency and accuracy of the resulting models.

\fakepar{Quantization} 
 Quantization for DL refers to using low-precision formats, such as INT8 and INT16, to represent the parameters of neural network models and intermediate activations, which are normally stored using a floating-point format. The quantization technique quantizes the float format values by multiplying them by a scale factor and rounding them to the nearest integer. Quantization can reduce the memory usage of both inference and training but at the cost of reduced accuracy of the resulting models. Researchers widely use this technique to design efficient ML algorithms. Mandheling, MiniLearn, TTE, and TinyTL are examples of works that apply quantization for on-device training optimization.

\fakepar{Dequantization} 
Dequantization is the process of converting quantized parameters of models and intermediate activations back to a high-precision format, which is the converse process of quantization. Quantization uses a scaling factor as a multiplier to map floating-point numbers into a smaller range of values, such as integers. Dequantization techniques normally represent the quantized values by dividing the scale factor to recover the original continuous values. By using dequantization, it is possible to mitigate the accuracy loss incurred during quantization, allowing for improved accuracy while still achieving the memory savings of quantization. MiniLearn combines quantization and dequantization techniques to enable on-device training.

\fakepar{Mixed-precision algorithms} 
 Researchers have proposed many mixed-precision algorithms~\cite{lin2017towards, lin2015neural, jacob2018quantization, courbariaux2015binaryconnect} for training, where the weights and intermediate activations are represented not only by FP32 but also by lower precision formats such as INT8 and INT16. The mixed-precision algorithm is the key technique used in Mandheling.

\fakepar{Pruning} 
 Pruning techniques can be broadly classified as network pruning and activation pruning~\cite{liu2018dynamic}. Network pruning removes less important units of an NN, while activation pruning builds a dynamic sparse computation graph to prune activations during training. MiniLearn uses network pruning as one of its main techniques.

\fakepar{Sparse update} 
 The sparse update refers to updating only a subset of the NN parameters during training while freezing the other parameters. For example, for a CNN model, TinyTL only updates the bias and freezes the parameters of filters. MiniLearn and TinyTL apply this strategy.

\fakepar{Continual Learning} 
 A typical ML pipeline can be simplified into three steps: model definition/selection, model training, and model inference. Various optimization techniques have been proposed to improve efficiency. For example, pruning techniques reduce the model size to enhance inference efficiency, while micro-batch techniques accelerate the training process. Unlike these techniques, continual learning enables DNN models to learn new tasks with a few data samples over a lifetime. As aforementioned, when applying continual learning, researchers need to address the catastrophic forgetting issue. The three common approaches are regularization approaches, dynamic architectures, and memory replay. The continuous growth technique is an example of the dynamic architecture approach. The technique optimizes the pipeline from the initial step. Rather than choosing and training a model with redundant parameters, the continuous growth approach starts with a small neural network and continuously increases its size during training until achieving a balance between accuracy and efficiency.

\fakepar{Adjusting the architecture of NNs} 
 TinyTL and TinyOL enable on-device training by modifying the architecture of NNs. TinyTL modifies the architecture by adding new layers to the network, while TinyOL designs an extra layer and then inserts it into the NN or uses the extra layer to replace one original layer of the NN. During training, both TinyTL and TinyOL only tune the additional layers to achieve high-efficient model training on resource-constrained devices.

\subsection{Auxiliary hardware}
Traditional model training leverages the GPU for its high parallel computation capability. When performing on-device training, there normally are no powerful GPUs, but auxiliary hardware is still available for workload offload. For example,  the DSPs are particularly suitable for integer operations and ubiquitously available on modern SoCs. Mandheling leverages a DSP to enable model training on smartphones.

\section{Performance Gains of On-device Training Systems}
\label{Sec:result}
This section presents the performance gains of the surveyed systems. Overall, researchers use one or several metrics from the following four metrics to measure their systems: accuracy improvement, memory usage reduction, energy consumption reduction, and training acceleration. Section~\ref{subsec:Baselines} summarises the baselines against which the systems compare their performance. Section~\ref{subsec:accuracy} describes the accuracy improvement brought by these on-device training methods. One of the goals of on-device training is to keep improving the accuracy of the deployed model. Therefore, this is the key outcome of on-device training. Section~\ref{subsec:memoryreduction} presents the optimized memory usage of the different systems.  Limited memory is the main bottleneck of on-device training systems. Hence, this is one of the most important metrics for evaluating the system. Section~\ref{subsec:energyconsumption} shows the energy consumption of model training, which is another concern for IoT devices that are normally battery-powered. Section~\ref{subsec:trainingspeed} describes the training speed improvement, which is another important evaluation metric for on-device training. The training tasks are expected to be completed within a reasonable time. Table~\ref{tab:performancegain} summarizes the performance gain of the discussed on-device training systems. 

\subsection{Baselines}
\label{subsec:Baselines}
 The baselines used for evaluation in current works mainly consist of four categories: Firstly, the ideal baseline, or cloud baseline. This baseline represents the situation with sufficient hardware resources, such as training models on the cloud or powerful servers, where sufficient memory and computing resources are provided. This baseline provides the best training accuracy and the corresponding resource consumption, such as memory. With this baseline, researchers show the potential accuracy gap and the reduction in resource consumption between on-device training and the ideal case. Secondly, training models without any optimization techniques or using pre-trained models without any further training. This baseline is used to show the performance gain of the proposed method, for example, accuracy improvement and memory consumption reduction. Thirdly, existing model training optimization techniques and frameworks,  this type of baseline is used to show the improvements of the proposed on-device training schemes, such as TensorFlow Lite~\cite{li2020tensorflow} and MNN~\cite{jiang2020mnn}, two efficient and lightweight DL frameworks that are commonly used for on-device training. Fourthly, baselines that only use one single optimization technique used in the system design. This aims to evaluate the impact of individual techniques on the results. 

\subsection{Accuracy Improvement}
\label{subsec:accuracy}
A primary objective of on-device training is to uphold or even enhance the accuracy of deployed models while operating under computational constraints. Consequently, accuracy improvement stands as a crucial performance metric for on-device training. Existing studies assess the accuracy gains achieved by on-device training systems across a spectrum of representative ML tasks encompassing computer vision (CV), audio recognition, and human activity recognition (HAR). Despite the utilization of diverse neural network architectures and benchmark datasets, we noticed that on-device training can contribute to accuracy enhancements. While the magnitude of accuracy gains may vary depending on the task, the findings collectively demonstrate the potential to achieve accuracy improvements without reliance on external servers or internet connectivity while also mitigating some critical privacy risks.  Furthermore, a notable advantage of on-device training is its capability to address the data drift problem. Data drift is a prominent challenge in ML systems after deployment~\cite{modeldrift}, which refers to the phenomenon where the distribution of the features changes after training, leading to decrement in the accuracy of models. 

\fakepar{Computer Vision (CV)} CV stands as a foundational domain within ML, underpinning critical advancements spanning from autonomous vehicles to automated diagnostics, thus establishing itself as one of the most impactful areas within the field. Evaluated CV tasks encompass image classification, object detection, and human facial attribute classification. These evaluations employ a diverse array of neural network architectures, including transformers and convolutional neural networks (CNNs), such as ImageNet~\cite{deng2009imagenet}, MobileNetV1~\cite{howard2017mobilenets}, MobileNetV2, SqueezeNet~\cite{iandola2016squeezenet}, ResNet-50 and Vision Transformer~\cite{dosovitskiy2020image}. Correspondingly, a wide spectrum of benchmark datasets is utilized, including Cars~\cite{krause20133d}, Flowers~\cite{nilsback2008automated}, Aircraft~\cite{maji2013fine}, CUB~\cite{wah2011caltech}, Pets~\cite{parkhi2012cats}, Food~\cite{bossard2014food}, CIFAR10~\cite{krizhevsky2009learning}, CIFAR100~\cite{krizhevsky2009learning}, VGGFace2~\cite{cao2018vggface2}, MiniImageNet~\cite{vinyals2016matching}, Tiny-ImageNet~\cite{le2015tiny}, ImageNet1k~\cite{russakovsky2015imagenet}, Stanford Dogs~\cite{khosla2011novel} and CelebA~\cite{liu2018large}.

In terms of accuracy enhancement, MiniLearn adopts a strategy of on-device model retraining with 600 data samples, resulting in significant accuracy improvements compared to pre-trained models and achieving comparable accuracy levels with the ideal baseline. Moreover, MiniLearn demonstrates the capability for a pruned model (with pruning percentages up to 75\%) to outperform pre-trained models and attain commendable accuracy levels with 400 to 500 samples. TinyTL, when applied to ImageNet models, achieves substantial accuracy improvements exceeding 30\% on benchmark datasets such as Cars and Aircraft, over 10\% on CIFAR100, Food, and CIFAR10, and enhancements ranging from 0.5\% to 5.4\% on datasets including Flowers, CUB, CelebA, and Pets. Furthermore, Melon conducts model training for other prominent CV architectures, including MobileNetV1, MobileNetV2, SqueezeNet, and ResNet-50, in both centralized and federated scenarios. In centralized settings, Melon achieves accuracy gains of 1.98\% and 2.04\% on MobileNetV2 and SqueezeNet, respectively. In federated scenarios, Melon achieves accuracy improvements of 3.94\% and 3.20\% over the baseline. For LifeLearner, the system achieves near-optimal accuracy of continual learning. In continual learning, the oracle baseline represents the optimal performance, because it can access all the classes in an i.i.d. manner and train the DNN model as many epochs as possible until converge. Compared to the Oracle baseline, LifeLearner only loses 0.2\% accuracy for CIFAR-100 dataset and 2.7\% accuracy for MiniImageNet dataset. In contrast, the previous SOTA Meta-CL method shows a 9.9\% accuracy drop for CIFAR-100 and 10.7\% for MiniImageNet compared to the Oracle baseline. Miro achieves 1.35–15.37\% higher accuracy across various memory budgets compared to the baseline systems with small to medium-scale datasets, and achieves 23.37\% higher accuracy compared to a SOTA HEM-based system CarM~\cite{lee2022carm} when using a larger dataset (ImageNet1k).

\fakepar{Audio Recognition} Audio recognition holds substantial importance within the ML domain, serving as a fundamental component in various sound-based intelligent applications, including voice-activated assistants. Harnessing the capabilities of on-device training, MiniLearn achieves significant accuracy gains in the Google Keyword Spotting (KWS) benchmark dataset~\cite{warden2018speech}, reaching comparable levels to the ideal baseline. 
LifeLearner shows a 5.6\% accuracy drop compared to the Oracle baseline for Google Speech Command V2 (GSCv2)~\cite{commands1804dataset}, while the previous SOTA Meta-CL method only reveals a 0.2\% accuracy reduction. However, the previous SOTA Meta-CL method is not suitable for resource-constrained devices. LifeLearner is an on-device continuous learning system that requires much lower system resources. Hence, the difference is acceptable.

\fakepar{Human Activity Recognition (HAR)} HAR plays an important role in enabling smart devices to intuitively comprehend and respond to human movements, thereby facilitating advancements in health monitoring, fitness tracking, and home automation. By enhancing technology's responsiveness and personalization to human activities, HAR contributes to improving the overall user experience. In a manner akin to audio recognition, MiniLearn leverages on-device training techniques to continually refine CNN models on-device, ultimately achieving accuracy levels on par with the ideal baseline in the WISDM-HAR benchmark dataset~\cite{kwapisz2011activity}. 

\subsection{Memory Usage Reduction}
\label{subsec:memoryreduction}
Memory scarcity poses a significant bottleneck in on-device training of DL models, particularly for high-constraint devices, which typically possess only kilobyte-level memory. While middle-constraint and low-constraint devices offer larger memory capacities compared to their high-constraint counterparts, the imperative to reduce memory usage remains paramount in training DNN models. Even server-scale GPUs, with their substantial memory resources, are grappling with escalating memory requirements for model training. Our review indicates the feasibility of on-device training across all device types. On-device training systems facilitate on-device training by adapting and refining well-established techniques utilized in server-based training, including micro-batching~\cite{huang2019gpipe} and recomputation~\cite{chen2016training}, alongside general memory management strategies like paging~\cite{tanenbaum2009modern}. Furthermore, for high-constraint devices, general efficient ML techniques such as pruning~\cite{molchanov2016pruning} and quantization~\cite{jacob2018quantization} play a pivotal role in memory optimization. Collectively, the diverse array of techniques aimed at enhancing memory efficiency holds immense potential in advancing on-device training capabilities.

\fakepar{High-constraint Devices} Several recent works, including MiniLearn, TTE, and POET, have demonstrated the feasibility of model training on high-constraint devices. In these studies, researchers either train simple NNs with fewer than ten layers or employ fine-tuning and transfer learning techniques to train more complex models. This approach is justified, as high-constrained devices lack the resources to support training from scratch.

For training simple NN models, MiniLearn conducts separate evaluations of Flash and RAM consumption. When retraining with 100 samples, MiniLearn achieves reductions of up to 58\%, 77\%, and 94\% in Flash consumption for KWS, CIFAR, and WISDM-HAR, respectively, compared to cloud-based training. With 600 samples, Flash consumption reduction rates are 8\% for KWS, 17\% for CIFAR, and 80\% for WISDM-HAR. Regarding RAM consumption, retraining without pruning consumes 196KB, 128KB, and 92KB on KWS, CIFAR, and WISDM-HAR, respectively. Fine-tuning only the fully connected layers reduces RAM consumption to 12KB, 10KB, and 6KB on KWS, CIFAR, and WISDM-HAR.

For complex model training, current systems employ fine-tuning to facilitate on-device training. In TTE, MobileNetV2, ProxylessNAs, and MCUNet are evaluated, using datasets such as Cars, CIFAR-10, CIFAR-100, CUB, Flowers, Food, and Pets. Peak memory usage is compared across three settings: fine-tuning the full model, sparse updates-only, and sparse updates with TTE graph reordering. Sparse updates effectively reduce peak memory usage by 7-9$\times$ compared to full fine-tuning while maintaining or exceeding transfer learning accuracy. Incorporating TTE graph reordering achieves total memory savings of up to 20-21$\times$. On the other hand, POET evaluates memory consumption for one training epoch on VGG16, ResNet-18, and BERT. Compared to the baseline POFO, POET reduces memory usage by 8.3\% and improves throughput by 13\%.

\fakepar{Middle-constraint Devices} Middle-constraint devices typically boast memory capacities ranging from megabytes to gigabytes but lack a dedicated GPU. Fine-tuning and transfer learning serve as viable strategies for enabling on-device training on these devices. For instance, TinyTL achieves memory savings of less than 20\% compared to the baseline by fine-tuning the normalization layers and the last layer, while delivering up to 7.2\% higher accuracy. Additionally, compared to fine-tuning the entire neural network, TinyTL achieves a 6$\times$ reduction in training memory while maintaining the same level of accuracy.

\fakepar{Low-constraint Devices} Low-constraint devices are typically robust devices equipped with several gigabytes of memory and GPUs. However, efficient memory management remains essential for training on such devices. In the case of Melon, the authors assess the impact of memory budget adaptation by training MobileNet V2 and SqueezeNet on a Samsung Note 10, with memory varying from 6GB to 5GB and from 4GB to 3GB, respectively. Melon achieves significant overhead memory savings ranging from 4.27\% to 54.5\% compared to a simple stop-restart approach, where the entire memory pool is reallocated, and all intermediate activations are discarded upon changes in memory budget. Another system, Sage, demonstrates memory usage reductions of up to 50\% compared to the DGA-only baseline~\cite{stein2021latency} when training BERT-small~\cite{devlin2018bert}, ResNet-50, DenseNet-121~\cite{huang2017densely}, and MobileNetV2 on smartphones. Additionally, Sage achieves peak memory consumption reductions of up to 420\% compared to the DGC-only baseline~\cite{kirisame2020dynamic}.

In the domain of continual learning systems, MDLdroidLite is evaluated by training LeNet, MobileNet, and VGG-11 models on four Personal Mobile Sensing (PMS) datasets (sEMG, MHEALTH, HAR, and FinDroidHR) and two image datasets (MNIST and CIFAR-10). While direct measurement of memory consumption reduction is not conducted, the evaluation focuses on model size reduction, which inherently leads to reduced memory consumption. Specifically, MDLdroidLite achieves parameter reductions of 28$\times$ to 50$\times$ and FLOPS reductions of 4$\times$ to 10$\times$ when training LeNet on PMS datasets. For MobileNet training, parameter and FLOPS reductions of 4$\times$ to 7$\times$ and 2$\times$ to 7$\times$, respectively, are observed. Although training a large-scale VGG-11 on-device proves costly and leads to battery drain, MDLdroidLite successfully trains a backbone VGG-11 model to achieve over 75\% accuracy with a minimal battery consumption of 1328mAh. 

\subsection{Energy Consumption Reduction}
\label{subsec:energyconsumption}
Minimizing energy consumption during ML model training is essential for fostering sustainable AI development, mitigating environmental impact, and cutting operational costs~\cite{an2023chatgpt}. This concern is particularly pertinent in the context of on-device training, where many devices rely on battery power and cannot sustain intensive energy-consuming tasks. Energy usage in model training primarily stems from two sources: computation and memory access. While advancements in processor technology enhance computing capabilities, they also exacerbate energy consumption issues. Similarly, while parallelism boosts computing performance, it also escalates energy consumption. Furthermore, memory access operations are significant contributors to energy consumption~\cite{Horowitz2014}. On-device training systems address these challenges by optimizing both computation and memory access during model training. For instance, Sage~\cite{Sage} achieves notable energy consumption reductions of up to 49.43\% compared to baselines, thanks to efficient memory management practices, all while maintaining or even improving accuracy.

\fakepar{High-constraint devices} High-constraint devices face limitations in computational resources and battery life, making faster training essential for optimizing their scarce resources. In MiniLearn, the authors conduct experiments across each use case 30 times, varying the number of data samples from 100 to 600. As expected, higher sample counts lead to increased energy consumption. For instance, with 100 samples, model retraining consumes $254 \pm 0.2 mJ$, $203 \pm 0.2 mJ$, and $138 \pm 0.2 mJ$ on KWS, CIFAR, and WISDM-HAR, respectively, while with 600 samples, these figures rise to $1486 \pm 0.1 mJ$, $1016 \pm 0.2 mJ$, and $786 \pm 0.1 mJ$. MiniLearn also demonstrates energy savings in communication compared to an ideal baseline, which involves uploading training samples to the cloud and downloading the retrained model. For instance, with 600 samples and BLE communications, energy consumption amounts to 45 mJ, 462 mJ, and 210 mJ for WISDM, CIFAR, and KWS, respectively.

\fakepar{Middle-constraint devices} Middle-constraint devices face similar energy consumption challenges as high-constraint devices. POET, an on-device training system designed for middle-constraint devices, demonstrates lower energy consumption compared to baselines. For instance, when training ResNet-18 on the Raspberry Pi 4, POET consumes only 41\% to 58\% of the energy compared to the Capuchin baseline~\cite{peng2020capuchin}. Compared to solutions relying solely on paging or rematerialization, POET achieves energy savings of up to 40\%.

\fakepar{Low-constraint devices} Low-constraint devices, such as smartphones, are expected to maintain usability over extended periods between charging cycles. Consequently, model training should not excessively drain battery life, potentially disrupting user experience by, for example, causing missed calls due to battery depletion.

Two-stage systems, Melon and Sage, are tailored for smartphone environments. Melon's evaluation involves training two neural networks, MobileNetV2 and SqueezeNet with a Batch Normalization layer, on a Meizu 16t smartphone. Melon consistently outperforms baselines, achieving energy savings ranging from 22\% to 49.43\%. In comparison to the ideal baseline, Melon exhibits an average energy consumption of only 11.4\% and as low as 2.1\% in the best-case scenario. Sage, on the other hand, is assessed by measuring energy usage per training iteration of ResNet-50 on a Samsung Galaxy S10. With each iteration consuming approximately 4.9J, a 1000-iteration training session only utilizes 8\% of the device's entire battery capacity, suggesting Sage as a suitable on-device training system for mobile devices.
In addition, ElasticTrainer also shows high energy efficiency. In contrast to training full DNN, ElasticTrainer achieves up to three-fold energy consumption reduction without noticeable accuracy loss.

For continual learning systems, the energy consumption evaluations of Mandheling span across various NN architectures, including VGG-11/16/19, ResNet-18/34, and Inception V3. Mandheling consistently outperforms baselines in per-batch training scenarios, achieving significant energy consumption reductions compared to competitors such as MNN-FP32 and TFlite-FP32. Notably, Mandheling demonstrates energy consumption reductions ranging from 3.21 to 11.2 times compared to MNN-FP32 and 2.01 to 12.5 times compared to TFlite-FP32. Additionally, in convergence scenarios, Mandheling exhibits substantial energy consumption reductions of 5.96 to 9.62 times in single-device settings and even greater reductions in distributed scenarios, reaching up to 13.1 times for a single client on datasets like FEMNIST and CIFAR-100.  For LifeLearner, compared to the previous SOTA Meta-CL method, the system reduces its energy consumption over 80.9\% and 94.2\%, for the CIFAR-100 and GSCv2 datasets on NVIDIA Jetson Nano, and 92\% and 96\% on Raspberry Pi 3+, respectively. For Miro, the authors report that its energy cost is always the lowest compared to baselines across benchmarks and memory budgets.

Finally, the reinforcement learning-based system zTT is evaluated on both Google Pixel 3a and NVIDIA JETSON TX2 platforms, covering a range of applications, including Aquarium, YOLO, Video rendering, Showroom VR, Skype, and Call of Duty 4. Compared to baseline methods such as Maestro and default DVFS methods, zTT demonstrates superior power consumption management while maintaining strict performance guarantees. Specifically, when utilizing JETSON TX2, zTT achieves a reduction in power consumption of 37.4\% and 23.9\% for FPS rates of 20 and 30, respectively, compared to the default method. Furthermore, zTT consistently meets target FPS requirements, a feat unattainable by the default method or Maestro. Although none of the techniques reach 60 FPS on Pixel 3a, zTT performs closest to the target FPS among the evaluated methods.

\subsection{Training Acceleration}
\label{subsec:trainingspeed}
Accelerating the training speed of ML models is crucial for enabling quicker experimentation and reducing both computational costs and energy use. Low-constraint devices, such as smartphones, are expected to run multiple processes to provide a good user experience. Meanwhile, they feature limited computational power and battery life. Quick model training boosts user experience by supporting online model updates and instant decision-making, independent of other functionalities. Training acceleration is another gain from on-device training. With efficiency optimization, on-device training systems markedly improve the training speed compared to baselines. In particular, Mandheling~\cite{Mandheling} increases the converge speed over 10$\times$. 

\fakepar{Low-constraint devices} Mandheling, Melon, MDLdroidLite, and ElasticTrainer involve the assessment of training acceleration. In Mandheling, the authors measure the training time of Mandheling on VGG-11~\cite{simonyan2014very}, ResNet-18~\cite{he2016deep}, LetNet~\cite{lecun1998gradient} and InceptionV3~\cite{szegedy2016rethinking}. Baselines are based on MNN~\cite{jiang2020mnn} and TFLite~\cite{li2020tensorflow}. Concretely, four baselines are used in the evaluation. (1) MNN-FP32: the Floating Point 32-bit-based training method of MNN. (2) MNN-INT: the  Integer 8-bit-based training method based on MNN. (3) MNN-FP32-GPU:  the Floating Point 32-bit-based training method on mobile GPU through the OpenCL backend. (4) TFLite-FP32:the Floating Point 32-bit-based-based training method provided by TFLite. In the single-device scenario, compared to MNN-FP32, Mandheling takes between $5.05\times$ to $6.27\times$ less time for convergence with a small loss in accuracy (1.9\% - 2.7\%). In contrast to MNN-INT8,  Mandheling reaches the same accuracy and spends $3.55\times$ less time. In a federated learning (FL) scenario, the authors measure Mandheling with training datasets FEMNIST~\cite{reddi2020adaptive} and CIFAR-100. In addition, they apply an FL simulation platform~\cite{yang2021characterizing} and two baselines that use the traditional FL protocol FedAvg~\cite{mcmahan2017communication}: FloatFL and Int8FL. Mandheling uses $5.25\times$ and $10.75\times$ less time to converge on FEMNIST and CIFAR-100, respectively.

For Melon, the authors measure the convergence speed with training throughput, which is defined as Equation~\ref{eq:throughput}. The throughput represents the number of data samples trained every second. The higher the throughput of the system, the higher the convergence speed. Overall, Melon always outperforms other baselines except for the ideal baseline, and it often achieves a similar throughput as the ideal baseline. In particular, the authors respectively evaluate Melon on NN with and without BN layers. For training NN without BN layers, Melon achieves $1.51\times$ - $3.49\times$ higher throughput than vDNN~\cite{rhu2016vdnn}. In contrast to Sublinear~\cite{chen2016training}, Melon reaches $13\times$ - $3.86\times$ higher throughput. And it also achieves $1.01\times$ - $4.01\times$ higher throughput than the Capuchin baseline.  For NN with BN layers, Melon almost reaches the same performance as the ideal baseline (below 1\%) and significantly outperforms the other baselines.

\begin{equation}
    throughput = \frac{BatchSize}{PerBatchLatency}
\label{eq:throughput}
\end{equation}

For MDLdroidLite, the authors compare its performance against three SOTA parameter adaption methods in CG: NeST-bridge, CGaP-select, and Net2WiderNet. Each experiment is conducted five times with training spanning 30 epochs. MDLdroidLite demonstrates notable speed improvements, achieving an average acceleration of 4.88$\times$, 2.84$\times$, and 3.12$\times$ compared to NeST-bridge, CGaP-select, and Net2WiderNet, respectively. Moreover, MDLdroidLite maintains superior convergence accuracy stability in contrast to the evaluated baselines.

For ElasticTrainer, when using the ResNet50 model and $\rho=50\%$ (2 times training speedup), ElasticTrainer can achieve similar accuracy in contrast to training the full model by training a much smaller model portion on both Jetson TX2 and Raspberry Pi 4B. Specifically, on the Pets and Stanford Dogs dataset, ElasticTrainer even achieves 1\%-2\% higher accuracy than training the full model. The reason is that training a smaller trainable DNN portion leads to less overfitting than training the full model. Comparatively, compared to training the full model, only training the last prediction module loses over 20\% accuracy on the CUB dataset, and training the last prediction module together with bias parameters and batch normalization layers loses over 10\% accuracy. Specifically, on Jetson TX2, ElasticTrainer can achieve a 2-3.4$\times$ speedup compared to training the full model. Its training speedup also outperforms up to 30\% compared to training the last prediction module, bias parameters, and batch normalization layers. Although only training the final prediction module brings 20\% more speedup, it provides the lowest accuracy.

\fakepar{Middle-constraint devices} 
The authors of ElasticTrainer also evaluated their system on a middle-constraint device, i.e., Raspberry Pi 4B.  Generally, as on Jetson TX2, ElasticTrainer achieves an accuracy similar to training the full model with less training time. Especially, ElasticTrainer outperforms training the full model on the Pets dataset by achieving 1\%-2\% higher accuracy. In contrast, only training the final prediction module, bias parameters, and batch normalization layers provides 25\% training speedup at the cost of 30\% accuracy loss.

\fakepar{High-constraint Devices} For MiniLearn and TTE, the authors evaluate the training speed of their systems. In MiniLearn, the authors repeat experiments on each use case 30 times and report the average training time using MiniLearn with a different number of data samples, from 100 to 600 samples. The more samples used, the more time cost. For example, when using 100 samples to retrain models, the training tasks $10 \pm 0.2 s$, $8 \pm 0.2 s$, and $6 \pm 0.2 s$ on KWS, CIFAR, and WISDM-HAR, respectively, while the time prolongs to  $56 \pm 0.1 s$, $40 \pm 0.2 s$ and $30 \pm 0.3 s$ when using 600 samples. Meanwhile, for TTE, the authors contrast three methods: fine-tuning the full model with TFLite Micro, sparse updates with TFLite Micro kernels, and sparse updates with TTE kernels. The sparse updates plus TTE kernels method increases the training speed by $23\times$ - $25\times$ times significantly compared to the sparse update with the TF-Lite Micro method. 

\begin{table}[p]
\caption{Performance Gains for the publications discussed in this work.}
\centering
\rotatebox{90}{
\begin{tabular}{|c|c|c|c|c|}
\hline
\textbf{Systems} & \textbf{Accuracy Improvement} & \textbf{Memory Usage} & \textbf{Energy Consumption} & \textbf{Training Speed} \\ \hline

\makecell{MiniLearn~\cite{minilearn}} & \makecell[tl]{ Increases accuracy for a\\ sub-set by 3\% to 9\% of \\the original DNN. }  & \makecell[tl]{Reduction of up to 50\% \\memory compared to \\the original DNN.} & \makecell[tl]{Consumes
138-254 mJ (100 \\training samples), 786-1016 \\ mJ (600 training samples).} & \makecell[tl]{Converges in
6s - 10s (100 \\training samples), 30s - \\56s (600 training samples).} \\ \hline

Sage~\cite{Sage} & \makecell{-}  & \makecell[tl]{Reduces memory use by \\ more than 20-fold \\compared to a baseline.} & \makecell[tl]{For each iteration, Sage \\consumes around 4.9J.} & \makecell{-} \\ \hline

\makecell{MDLdroidLite\\ \cite{zhang2020mdldroidlite}}& \makecell{-} & \makecell[tl]{Achieves $28\times$ to $50\times$ \\parameter reduction\\ and $4\times$ to $10\times$ FLOPS \\reduction.} & \makecell{-} & \makecell[tl]{Speed up training by up to\\ $4.88\times$.} \\ \hline

Melon~\cite{Melon} & \makecell[tl]{Accuracy improvement, in\\ centralized training, 2.04\%, \\in federated training, 3.94\%,\\compared to MNN~\cite{jiang2020mnn}. } & \makecell[tl]{Achieves up to $4.33\times$ \\larger batch size under\\ the same memory\\budget.} & \makecell[tl]{Saves up to 49.43\% energy \\compared to baseline.} & \makecell[tl]{Achieves $1.89\times$ on average \\(up to $4.01\times$) higher \\training throughput.} \\ \hline

 \makecell{Mandheling\\~\cite{Mandheling}}  & \makecell{-} &  \makecell{-} & \makecell[tl]{Consumes $5.96-9.62\times$ less\\ energy for convergence.} & \makecell[tl]{Takes up to $6.27\times$ less time \\for convergence.} \\ \hline

TinyTL~\cite{cai2020tinytl} & \makecell[tl]{Accuracy improvement \\up to 34.1\%.} & \makecell[tl]{Saves the memory \\up to 6.5×. }& \makecell{-} & \makecell{-} \\ \hline

TTE~\cite{lin2022device} & \makecell{-} & \makecell[tl]{Reduces peak memory \\usage by $20\times$ - $21\times$. } & \makecell{-}  & \makecell[tl]{Increases the speed by\\ $23\times$ - $25\times$. } \\ \hline

POET~\cite{patil2022poet} & \makecell{-} & \makecell[tl]{Reduces memory usage\\
by 8.3\%.} & \makecell[tl]{Saves up to 59\% energy \\compared to baselines. } &  \makecell[tl]{Improves the throughput\\ by 13\%.} \\ \hline

zTT~\cite{kim2021ztt} & \makecell{-} & \makecell{-} & \makecell[tl]{23.9\% less average power \\consumption.} & \makecell{-} \\ \hline

LifeLearner~\cite{kwon2023lifelearner} &  \makecell[tl]{Accuracy improvement \\up to 8\% compared to \\the SOTA.} & \makecell[tl]{Saves the memory up \\to $178.7\times$ compared to\\ the SOTA.} & \makecell[tl]{Saves the energy up \\to 96\% compared to\\ the SOTA.} & \makecell{-} \\ \hline

ElasticTrainer~\cite{huang2023elastictrainer} & \makecell[tl]{Accuracy improvement \\ 1\%-2\% compared to \\full training.} & \makecell[tl]{Consumes 10\% more \\ memory than full \\training.} & \makecell[tl]{Saves the energy up to \\ $3\times$ compared to full \\ training.} & \makecell[tl]{Increases the speed by\\  $3.4\times$ compared to full \\ training.} \\ 

\hline
\end{tabular}
}
\label{tab:performancegain}
\end{table}

\section{Summary of State-of-the-art Solutions}
\label{sec:discussion}
In this section, we elaborate on the methods and ideas presented in the current SOTA systems for on-device training (Section~\ref{Sec:systems}). These works target and address systematic issues raised by devices' resource limitations via different approaches or dimensions. Overall, the core mechanisms presented in SOTA works can be summarized in the following four categories.

\fakepar{Run-time optimization}
 This approach refers to applying various techniques to reduce resource consumption during training, which consists of two directions. The first one is using memory consumption optimization techniques to reduce peak memory consumption. Researchers exploring this direction of research conclude that memory is the main bottleneck for on-device training due to backpropagation causing a large number of intermediate variables. Hence, several works, including POET, Melon, and Sage incorporate memory optimization techniques, such as paging and materialization (Section~\ref{subsec:twostagesystem}). The second direction is to reduce resource consumption from the model perspective. For example, model quantization techniques can significantly reduce the size of the model and, as a result, decrease the resource requirement.  However, quantized models naively are more challenging to train due to precision reduction ( e.g., from float32 to int8). Minilearn and TTE, respectively, present solutions for this problem to allow for effective on-device training (Section~\ref{subsec:onestagesystem} and Section~\ref{subsec:twostagesystem}). In addition, selectively training the parameters of the model can also reduce the training time and resource consumption. Unfortunately, this leads to limited generalization ability. TinyTL addresses this issue by designing a lighter residual model (Section~\ref{subsec:onestagesystem}). 
Overall, run-time optimization is the most common strategy for on-device training. It helps on-device training systems achieve efficient memory usage (Section~\ref{subsec:memoryreduction}), which represents one of the most significant challenges of on-device training. In addition, it can also reduce computational complexity with optimization techniques from the model perspective, such as model quantization techniques.

\fakepar{Workload offloading} 
 Leveraging other available hardware is another possibility. For example, the authors of Mandheling observe that DSPs are particularly suitable for integer operation and ubiquitously available on modern smartphones. Hence, they apply CPU-DSP co-scheduling with a mixed-precision training algorithm to enable on-device training (Section~\ref{subsec:mandheling}). 
Even though auxiliary hardware is not commonly available for high-constraint devices, systems designed for low-constraint devices, such as smartphones, can gain both energy consumption reduction and training acceleration by offloading workload to auxiliary hardware (Section~\ref{subsec:energyconsumption} and Section~\ref{subsec:trainingspeed}). 

\fakepar{Continuous growth} 
 This approach tries to reduce resource consumption from the very beginning with the understanding that massive resource consumption is caused by the large size of NN models. Given a target NN, Frankle et al.~\cite{frankle2018lottery} show that traditional DNN models often have a large number of redundant parameters that can be reduced with little accuracy deduction. Based on this observation, MDLdroidLite starts from training a tiny NN and gradually grows the architecture until the accuracy satisfies the application's requirements (Section~\ref{subsec:MDLdroidLite}).
Continuous growth significantly reduces the number of parameters by up to 50 times and FLOPs by up to 10 times. In addition, it speeds up training by up to 4.88 times (Section~\ref{subsec:memoryreduction} and Section~\ref{subsec:trainingspeed}). Hence, this is one feasible mechanism for enabling on-device training.

\fakepar{Reinforcement learning} 
 RL is another approach to enable on-device training, where an agent interacts with an environment to learn a behavior or task through trial and error. The agent in the system learns to maximize the cumulative reward provided by the environment by selecting different possible actions. This paradigm implies continuous learning during runtime and does not require newly collected and labeled data. For example, zTT applies RL to design and implement a learning-based DVFS solution on mobile devices (Section~\ref{subsec:ztt}).

 As discussed above, from the ML perspective, current systems cover ML model-side optimization, training time optimization, and choosing ML methods with suitable paradigms (e.g., reinforcement learning). From the device side, most system designers configure their target systems using devices with sufficient computational capability (e.g., smartphones), which offers them flexibility in system design (e.g., GB-level memory and dedicated hardware such as DSP), as shown in Table~\ref{tab:device}. An additional benefit of using such commercially available (relatively) powerful platforms is the fact that they operate on accessible OSes (e.g., Android OS), providing a friendly development environment to migrate and distribute the scheme to many devices. Therefore, one possible research direction is to combine different approaches to achieve a more efficient hybrid solution. For example, one can combine continuous growth or RL and optimized memory management techniques to enable on-device training. Given that many of these approaches are orthogonal to each other and can be used together, understanding and combining the benefits of each approach can be an attractive future research direction.

\section{Conclusion and Future Directions}
\label{sec:conclusion}
 In this work, we present a survey of existing systems for on-device training. Although on-device learning is still an emerging field of research, a number of systems have been introduced to support general on-device training for mobile and IoT platforms. Mobile platforms play a crucial role in people's daily lives and have sufficient hardware capabilities to apply various optimization techniques for model training, including several gigabytes of memory, powerful CPUs, and hardware accelerators such as GPUs, NPUs, and DSPs. On-device training, therefore, can facilitate the development of intelligent applications for smartphones and migrate the integration from the cloud infrastructure to the devices themselves. On the other hand, for many IoT platforms, which are microcontroller-based and hold tighter resource limitations, model training becomes more challenging due to their extreme resource constraints. Hence, on-device training systems for microcontroller platforms typically require both model-side and resource-consumption optimizations. Overall, on-device learning allows various devices to locally leverage the power of ML, and we anticipate it to become an important aspect of many applications for industry and society.

 At the same time, our survey also leaves us with a list of future research directions, that are yet to be deeply explored. Specifically, fewer efforts have been made for enabling on-device model training for highly constrained IoT devices, such as those that feature only a couple hundred MBs of memory and clock speeds in the sub-GHz range. Despite their significance and adaptation in designing a wide range of smart applications, the lack of systems tailored for such platforms is mainly due to two reasons. Firstly, such constrained hardware gives less flexibility for system design. Secondly, there is no one-size-fits-all OS for these IoT devices; thus, requiring per-system/implementation optimization. Designing a common framework that enables per-platform customized on-device learning for resource-constraint IoT platforms can be an interesting direction for future research. Potentially, we can think of approaches such as (i) exploring the possibility of applying existing optimized techniques on tiny devices, (ii) designing specific efficient NN for tiny devices with considerations on both inference and training efficiency, and (iii) integrating ML features, both inference and training, on the OS level to ease future development and migration. Future research can be initiated by exploiting mature embedded OSes, such as Contiki~\cite{dunkels2004contiki}, TinyOS~\cite{levis2005tinyos}, FreeRTOS~\cite{barry2008freertos}, or Android Things~\cite{roy2020android}.

\section*{Acknowledgements}
This work was partially supported by the Swedish Science Foundation (SSF), Korean Ministry of Science and ICT (MSIT) and the IITP under grants IITP-2024-2020-0-01461 (ITRC Program) and IITP-2022-0-00420.

\bibliographystyle{ACM-Reference-Format}
\bibliography{sample-base}

\end{document}